\documentclass[10pt,journal,compsoc]{IEEEtran}
\ifCLASSOPTIONcompsoc
  \usepackage[nocompress]{cite}
\else
  \usepackage{cite}
\fi
\ifCLASSINFOpdf
\else
\fi
\usepackage{booktabs} 
\usepackage{subfig}
\usepackage{graphicx}
\usepackage{algorithm}
\usepackage{algpseudocode}%
\usepackage{xcolor}
\usepackage{amsmath}
\usepackage{comment}
\newcommand{\etal}{\textit{et al}. }
\usepackage{mathtools}
\usepackage{multirow}
\usepackage{graphics}
\usepackage{enumitem}
\usepackage{CJKutf8}
\usepackage{array}
\usepackage{diagbox}
\usepackage{hyperref}
\usepackage[utf8]{inputenc}
\usepackage{bm}

\usepackage{ragged2e}
\usepackage{amsmath,amssymb}

\usepackage{xcolor,cite,etoolbox}
\definecolor{mypurple}{rgb}{0.4392, 0.1882, 0.6275}
\definecolor{mygreen}{rgb}{0, 0.6902, 0.3137}
\makeatletter 
\pretocmd\@bibitem{\csname keycolor#1\endcsname}{}{\fail}
\newcommand\citecolor[1]{\@namedef{keycolor#1}{}}
\renewcommand{\algorithmicrequire}{\textbf{Input:}}

\begin{document}
%
\title{CE-SSL: Computation-Efficient Semi-Supervised Learning for ECG-based Cardiovascular Diseases Detection}
%
%
%
%
\author{Rushuang Zhou, Lei Clifton, Zijun Liu , Kannie W. Y. Chan, \\ David A. Clifton, Yuan-Ting Zhang,~\IEEEmembership{Fellow,~IEEE} and Yining Dong
\IEEEcompsocitemizethanks{\IEEEcompsocthanksitem Rushuang Zhou, Zijun Liu, Kannie W.Y. Chan are with the Department
of Biomedical Engineering, City University of Hong Kong, Hong Kong,
China and also with Hong Kong Center for Cerebro-Cardiovascular Health Engineering (COCHE), Hong Kong, China.
\IEEEcompsocthanksitem Kannie W.Y. Chan is also with the Russell H. Morgan Department of Radiology and Radiological Science, The Johns Hopkins University School of Medicine, Baltimore, Maryland, USA, and also with the Shenzhen Research Institute, City University of Hong Kong, Shenzhen, China. 
\IEEEcompsocthanksitem Lei. Clifton is with the Nuffield Department of Population Health, University of Oxford, UK.
\IEEEcompsocthanksitem David A. Clifton is with the Department of Engineering Science, University of Oxford, UK, and also with Oxford-Suzhou Institute of Advanced Research (OSCAR), Suzhou, China.
\IEEEcompsocthanksitem Yuan-Ting Zhang is with the Department
of Biomedical Engineering, Chinese University of Hong Kong, Hong Kong,
China and also with the Hong Kong Institutes of Medical Engineering, Hong Kong, China.
\IEEEcompsocthanksitem Yining Dong is with the Department
of Data Science, City University of Hong Kong, Hong Kong,
China and also with Hong Kong Center for Cerebro-Cardiovascular Health Engineering (COCHE),  Hong Kong, China.
\IEEEcompsocthanksitem Corresponding authors: Yining Dong, e-mail: yinidong@cityu.edu.hk.}

\thanks{Manuscript received April 19, 2005; revised August 26, 2015.}}

%
%

\markboth{Journal of \LaTeX\ Class Files,~Vol.~14, No.~8, August~2015}%
{Shell \MakeLowercase{\textit{et al.}}: Bare Demo of IEEEtran.cls for Computer Society Journals}
%



\IEEEtitleabstractindextext{%
\begin{abstract}
The label scarcity problem is the main challenge that hinders the wide application of deep learning systems in automatic cardiovascular diseases (CVDs) detection using electrocardiography (ECG). Tuning pre-trained models alleviates this problem by transferring knowledge learned from large datasets to downstream small datasets. However, bottlenecks in computational efficiency and detection performance limit its clinical applications. It is difficult to improve the detection performance without significantly sacrificing the computational efficiency during model training. Here, we propose a computation-efficient semi-supervised learning paradigm (CE-SSL) for robust and computation-efficient CVDs detection using ECG. It enables a robust adaptation of pre-trained models on downstream datasets with limited supervision and high computational efficiency. First, a random-deactivation technique is developed to achieve robust and fast low-rank adaptation of pre-trained weights. Subsequently, we propose a one-shot rank allocation module to determine the optimal ranks for the update matrices of the pre-trained weights. Finally, a lightweight semi-supervised learning pipeline is introduced to enhance model performance by leveraging labeled and unlabeled data with high computational efficiency. Extensive experiments on four downstream datasets demonstrate that CE-SSL not only outperforms the state-of-the-art methods in multi-label CVDs detection but also consumes fewer GPU footprints, training time, and parameter storage space. As such, this paradigm provides an effective solution for achieving high computational efficiency and robust detection performance in the clinical applications of pre-trained models under limited supervision. Code and Supplementary Materials are available at \textit{https://github.com/KAZABANA/CE-SSL }
\end{abstract}

\begin{IEEEkeywords}
Electrocardiograph; Semi-Supervised Learning; Cardiovascular Diseases; Computation-Efficient.
\end{IEEEkeywords}}

\maketitle

\IEEEdisplaynontitleabstractindextext

%
\IEEEpeerreviewmaketitle

\IEEEraisesectionheading{\section{Introduction}\label{sec:introduction}}

%
%
%
%
\IEEEPARstart{C}ardiovascular diseases have become the deadliest 'killer' of human health in recent years\cite{kelly2010promoting}. As a non-invasive and low-cost tool, ECG provides a visual representation of the electrical activity of the heart and is widely used in the detection of various CVDs\cite{kiyasseh2021clinical,lai2023practical}.  Benefiting from recent progress in computing hardware, ECG-based deep learning systems have achieved notable success in automatic CVDs detection\cite{hannun2019cardiologist,ribeiro2020automatic,al2023machine,lu2024decoding}. However, previous deep learning models required sufficient labeled samples to achieve satisfactory performance when trained on new application scenarios with unseen CVDs\cite{berthelot2019mixmatch,sohn2020fixmatch}. Unfortunately, collecting well-labeled ECG recordings requires physicians' expertise and their laborious manual annotation,  and therefore is expensive and time-consuming in clinical practice\cite{zhang2022semi,zhoupami2023}. Recent advancements in pre-trained models have enhanced the performance of deep learning models on the downstream datasets without large-scale labeled data\cite{vaswani2017attention,radford2019language,he2022masked}. A commonly used pipeline consists of pre-training over-parameterized backbone models on large-scale datasets and then fine-tuning them on small downstream datasets in a supervised manner. However, two bottlenecks still greatly limit the clinical application of CVDs detection systems based on pre-trained models under limited supervision. 

\textbf{(1) The bottleneck in CVDs detection performance.} Fine-tuning of pre-trained models is currently conducted in a purely supervised manner. When the labeled data is very scarce in the downstream datasets, model performance may drop due to over-fitting\cite{selftune2021}. Fortunately, a large number of unlabeled data in the medical domain are relatively easy to collect. Semi-supervised learning (SSL) is able to extract sufficient information from the unlabeled data and outperform the supervised models trained with the same amount of labeled data\cite{zhou2018semi,sohn2020fixmatch,comatch2021,zhang2021flexmatch,peiris2023uncertainty}. For example, self-tuning integrates the exploration of unlabeled data and the knowledge transfer of pre-trained models into a united framework, which significantly outperforms supervised fine-tuning on five downstream tasks\cite{selftune2021}. Despite their robust performance, existing SSL methods are mainly built on pseudo-label techniques and the weak-strong consistency training on unlabeled samples\cite{berthelot2019mixmatch,berthelot2019remixmatch,sohn2020fixmatch,zhang2021flexmatch,chen2023softmatch}, which greatly increase the GPU memory footprint and computation time during model training. This drawback results in the bottleneck of computational efficiency during the performance enhancement of pre-trained models using semi-supervised learning.

\textbf{(2) The bottleneck in computational efficiency for parameter optimization.} Nowadays, many studies have introduced large-scale foundation models to achieve better CVDs detection performance using ECG\cite{vaid2023foundational,han2024foundation,mathew2024foundation,mckeen2024ecg, pham2024c}, greatly increasing the computation costs of modifying them for downstream applications. SSL methods and fine-tuning both update all the model parameters. Despite their effectiveness, both methods have the main drawback that they require saving the gradients of all the parameters and even the momentum parameters, resulting in large GPU memory footprints when tuning large pre-trained models\cite{hu2022lora}. Additionally, each tuned model can be regarded as a full copy of the original models, therefore leading to high storage consumption when simultaneously tuned on multiple datasets \cite{zhang2023adaptive}. To address this, parameter-efficient fine-tuning (PEFT) methods have been introduced to reduce the trainable parameters during model training and thus decrease the computational costs during model training\cite{houlsby2019parameter,zaken2021bitfit,chen2023hadamard}. For example, Low-rank adaptation (LoRA) achieves this goal by updating the pre-trained weights with low-rank decomposition matrices. AdaLoRA and IncreLoRA overcome the performance bottleneck of LoRA by allocating different ranks to different pre-trained weights based on their importance\cite{zhang2023adaptive,zhang2023increlora}. However, the above performance improvement is achieved at the cost of increased training time for iterative importance estimation. 

Therefore, a dilemma is encountered:  model performance improvement often comes at the expense of a large sacrifice of computational efficiency during model training. Specifically, semi-supervised learning enhances CVDs detection performance under limited supervision but at significantly increased computational costs. Conversely, methods that prioritize computational efficiency may compromise model performance\cite{ding2023parameter}. Consequently, achieving a superior detection performance with high computation efficiency poses a great challenge to the clinical application of pre-trained models in ECG-based CVDs detection. To the best of our knowledge, no prior study has designed and evaluated a framework to escape the dilemma.

Here, we propose a united paradigm capable of addressing the above two bottlenecks simultaneously. It is a computation-efficient semi-supervised learning paradigm (CE-SSL) for adapting pre-trained models on downstream datasets with high computational efficiency under limited supervision. Our method enables robust and low-cost detection of CVDs in clinical practice using ECG recordings. As shown in Fig.\ref{fig:flowchart}, first, a base backbone is pre-trained on a large-scale 12-lead ECG dataset in a supervised manner, which provides a foundation for downstream datasets. We also provide medium and large backbones for performance enhancement by increasing the backbone's depth and width. Second, a random-deactivation low-rank adaptation (RD-LoRA) method formulates a low-cost and robust pipeline for updating the pre-trained backbone on downstream datasets. Specifically, it stochastically activates or deactivates low-rank adaptation in each trainable layer of the backbone with a probability $p$. To reduce GPU memory footprint, the pre-trained weights in each layer are always frozen.  Theoretical analysis indicates that the random deactivation operation integrates various sub-networks generated during model training, thus overcoming the performance bottleneck in tuning pre-trained models. Additionally, deactivating low-rank adaptation in some layers reduces computation costs and speeds up the training process, especially when the backbone model size is large. Third, a one-shot rank allocation module allocates the optimal ranks for the low-rank matrices in each layer. In contrast to AdaLoRA\cite{zhang2023adaptive} and IncreLoRA\cite{zhang2023increlora}, the proposed method can determine the optimal ranks using only one gradient backward iteration, improving the adaptation performance at low computational costs. 

Additionally, a lightweight semi-supervised learning module is developed to leverage the abundant information within unlabeled data. This module uses unlabeled data to stabilize the statistics estimation process in batch normalization layers, enhancing their generalization performance on unseen data distributions. Compared to the pseudo-labeling and the weak-strong consistency training methods\cite{berthelot2019remixmatch,sohn2020fixmatch,chen2023softmatch}, our lightweight module alleviates the label scarcity problem with significantly higher computational efficiency. 

Finally, extensive experiments on four downstream datasets demonstrate the superior CVDs detection performance of the proposed CE-SSL against various state-of-the-art models under very limited supervision. Most importantly, our method only requires 66.5\% training time, 70.7\% GPU memory footprint, and 1.8\%-5.8\% trainable parameters of the state-of-the-art SSL methods. Furthermore, its computational costs can be minimized to adapt to resource-limited environments without a significant accuracy loss.  In conclusion, our proposed computation-efficient semi-supervised learning paradigm provides an effective solution to overcome the two bottlenecks that limit the clinical applications of pre-trained models in ECG-based CVDs detection. We summarize the major contributions as follows:
\begin{itemize}
\item Pre-trained backbones with various sizes are provided to serve as a foundation for ECG-based CVDs detection on downstream datasets. 
\item A random deactivation low-rank adaptation method is proposed to update the backbones with high computational efficiency and robust performance. A one-shot rank allocation module is present to determine the optimal rank distribution during low-rank adaptation at minimal costs.
\item A lightweight semi-supervised method is proposed to leverage large-scale unlabeled data without greatly sacrificing the computational efficiency. 
\item A computation-efficient semi-supervised framework for low-cost and accurate CVDs detection is proposed, which is the first one to escape the dilemma between model performance and computational efficiency.
\end{itemize}
\section{Related Work}
\label{sec:relatedWorksemi}
\subsection{AI-Enabled CVDs Prediction using ECG }
Benefiting from the development of deep learning, AI-enabled systems have shed light on automatic ECG screening and cardiovascular disease diagnosis\cite{Pour2018, hannun2019cardiologist,ribeiro2020automatic,strodthoff2020deep,kiyasseh2021clocs,huang2022snippet,vaid2023foundational,han2024foundation,mathew2024foundation}. Tracing the development of the systems, it can be observed that the prediction models they used are continuously scaling up. In the first stage, small-scale models demonstrated promising diagnosis performance in ECG analysis and CVDs detection. For example,  Pourbabaee \etal designed a deep convolutional neural network to extract features from ECG signals and utilize standard classifiers for screening paroxysmal
atrial fibrillation\cite{Pour2018}. Hannun \etal proposed an end-to-end deep convolutional neural network to achieve automatic single-lead ECG screening\cite{hannun2019cardiologist}. The results demonstrated that the network achieved similar diagnosis performance compared with common cardiologists. In the second stage, pre-trained models with a prohibitive number of parameters were introduced, which demonstrated better transferability than previous networks. This advantage reduces their requirement for supervision information on downstream datasets. For instance, Vaid \etal pre-trained a large-scale vision transformer (HeartBEiT) on a huge ECG dataset and fine-tuned it on downstream datasets. The experiment results demonstrated the superiority of HeartBEiT in CVDs detection compared with traditional CNN architectures. In the current stage, many studies have proposed various kinds of foundation models for more advanced ECG screening and cardiac healthcare, inspired by their success in natural language processing\cite{han2024foundation,mathew2024foundation}. However, pre-trained models might experience a performance drop on downstream datasets when the labeled samples are very scarce there. Additionally, the computational costs of adapting them to various tasks significantly increase as their sizes scale up.

\subsection{Semi-Supervised Learning for Performance Enhancement under Limited Supervision.}
Semi-supervised learning offers an effective solution to address the label scarcity problem by leveraging unlabeled samples\cite{selftune2021, berthelot2019mixmatch, sohn2020fixmatch, zhang2021flexmatch, chen2023softmatch}.  For example, Sohn \etal combined consistency regularization and pseudo-labeling to formulate a powerful algorithm (FixMatch)\cite{sohn2020fixmatch}. Extensive experiments demonstrate the superiority of FixMatch against the supervised baselines under the label scarcity condition. Subsequently, Zhang \etal proposed curriculum pseudo labeling (CPL) to flexibly adjust the thresholds for pseudo label selection, aiming at utilizing unlabeled data based on the model's training progress\cite{zhang2021flexmatch}. Using a truncated Gaussian function, Chen \etal designed a soft threshold to weight unlabeled samples according to their prediction confidence, which achieved a balance between pseudo labels quality and quantity\cite{chen2023softmatch}. Compared with FixMatch, FlexMatch and SoftMatch both demonstrate better performance in various datasets. However, Wang \etal pointed out that the performance of semi-supervised models will be influenced by inaccurate pseudo labels, especially in large label space\cite{selftune2021}. Hence, they proposed a self-tuning technique to explore the potential of the transfer of pre-trained models and a pseudo-label group contrast mechanism to increase the model's tolerance to inaccurate labels. Experiments on five tasks demonstrated the superiority of the proposed framework against previous semi-supervised and supervised methods. In summary, massive unlabeled data and powerful pre-trained models led to the success of semi-supervised methods. However, high computation burdens are the side effects of leveraging them, greatly limiting their applications in resource-limited settings.

\subsection{Parameter-Effcient Methods for Higher Computational Efficiency.} 
Parameter-efficient training has demonstrated great potential in decreasing the computational costs of fine-tuning pre-trained models\cite{zaken2021bitfit,hu2022lora,2023dylora,zhang2023adaptive}. For example, Zaken \etal proposed BitFit to fine-tune the bias terms of the pre-trained models and freeze the other parameters, greatly reducing the computational costs. However, BitFit sacrifices the performance of the fine-tuned models because most of their parameters are not well adapted to downstream tasks. Hu \etal designed a low-rank adaptation method (LoRA) to inject trainable low-rank matrices into the transformer architecture, decreasing the performance gap between parameter-efficient methods and full fine-tuning\cite{hu2022lora}. However, Zhang \etal pointed out that LoRA ignored the varying importance of different pre-trained weights and allocated the same rank for all the trainable matrices, which led to suboptimal fine-tuning performance\cite{zhang2023adaptive}. Consequently, they designed AdaLoRA to address this problem, which dynamically allocates different ranks to the low-rank matrices according to their importance during fine-tuning. During this process, the trainable parameters of the matrices with low importance are pruned.  Different from AdaLoRA, IncreLoRA adaptively adds trainable parameters to the low-rank matrices with high importance\cite{zhang2023increlora}. As a non-pruning method, its performance is not limited by the preset parameter budget. Although IncreLoRA and AdaLoRA surpass LoRA in some scenarios, they result in high computation costs for weight importance estimation. Consequently, advancing fine-tuning performance without sacrificing computational efficiency remains challenging when designing parameter-efficient methods. 

\begin{figure*}[h]
\begin{center}
\includegraphics[width=1\textwidth]{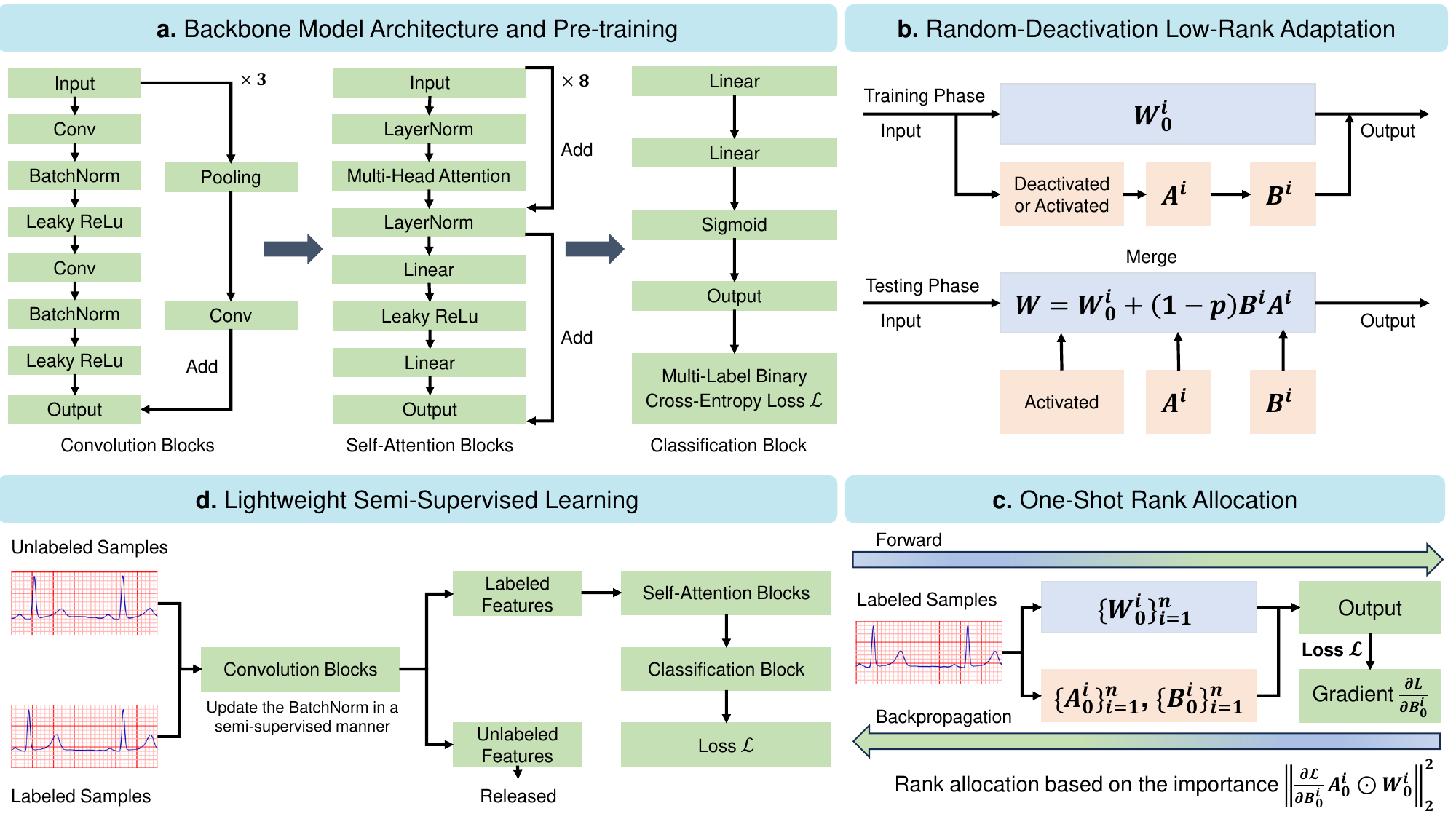}
\end{center}
\caption{\textbf{Overview of CE-SSL.} \textbf{a. The architecture of the pre-trained backbone model.} It consists of three convolution blocks, eight self-attention blocks, and one classification block in the base backbone. It is pre-trained on a public 12-lead ECG dataset using the supervised multi-label binary cross-entropy loss. \textbf{b. Random-deactivation low-rank adaptation.} On the downstream datasets, the pre-trained weights $\{W_{0}^{i}\}_{i=1}^{n}$ in the backbone are updated by the proposed random-deactivation low-rank adaptation. It randomly activates or deactivates the low-rank matrices ($\{B^{i}\}_{i=1}^{n}$ and $\{A^{i}\}_{i=1}^{n}$) in each trainable layer with a given probability $p$. Note that all the pre-trained weights are frozen during model training. All the low-rank matrices are activated and merged into the pre-trained weights in the testing stage. The merge process generates an ensemble network combining all the sub-networks produced by the random deactivation operation. \textbf{c. One-shot rank allocation.} The ranks of the low-rank matrices are determined by the proposed one-shot rank allocation method using only one gradient backward on the labeled samples. The matrices with high importance are allocated with a higher rank than those with low importance. \textbf{d. Lightweight semi-supervised learning.} During the low-rank adaptation process, unlabeled samples from the downstream datasets are combined with the labeled samples to estimate the statistics in batch-normalization layers. Subsequently, only the labeled data is forwarded to the self-attention blocks for CVDs detection, and the unlabeled data is released in GPU memory. This lightweight semi-supervised pipeline improves the model performance in a computational-efficient way.  } 
\label{fig:flowchart}
\end{figure*}

\begin{table*}[h]
\begin{center}
\caption{Backbone model specifications. $N_{conv}$ indicates the number of convolution blocks, $N_{att}$ indicates the number of self-attention blocks, and $N_{cls}$ indicates the number of classification blocks. $C$ is the number of convolution channels. Hidden size is the hidden layer dimension of the self-attention blocks. Head Num is the number of heads in multi-head self-attention. Params is the total number of parameters in the backbone.}
\label{tab:settings}
\scalebox{1}{
\color{black}
\begin{tabular*}{\hsize}{@{}@{\extracolsep{\fill}}lccccccc@{}}
\toprule
Backbone Size & $N_{conv}$ & $N_{att}$ & $N_{cls}$ & $C$ & Hidden size & Head Num & Params\\
\midrule
Base &3 &8 & 1 & 256 & 256&16&9.505M\\
Medium &3 &12 & 1 & 512 & 512 &16&50.494M\\
Large &3 &12 & 1 & 768 & 768&16&113.490M\\
\bottomrule
\end{tabular*}
}
\end{center}
\end{table*}

\section{Methodology}
\label{sec:methodology}
\subsection{Backbone Model Pre-Training}
The base backbone model is pre-trained on a public 12-lead ECG dataset (CODE-15\%\cite{ribeiro2019tele,ribeiro2020automatic}), where 345779 ECG recordings from 233770 patients are provided. The medium and large backbones are pre-trained on a restricted dataset with 2,322,513 ECG recordings from 1,558,772 patients (CODE-full\cite{ribeiro2019tele,Leilu2024}). The specific settings of the backbone models with different sizes are shown in Table \ref{tab:settings}. Note that multiple abnormalities could be identified from one ECG recording simultaneously, which indicates that a multi-label classification model should be implemented for ECG-based CVDs detection. As shown in Fig.\ref{fig:flowchart}, The backbone model $M(X)$ consists of three parts: (1) Convolution blocks, (2) Self-attention blocks, and (3) Classification blocks. Specifically, the convolution blocks comprise multiple convolution layers (Conv) and batch normalization layers. The Leaky-Relu function is used as the activation function and skip-connection is implemented\cite{AIMTbackbone}. In addition, a simple but efficient self-attention pipeline is employed in the self-attention blocks\cite{radford2019language} and two successive fully-connected layers with sigmoid activation are used for label prediction in the classification block.  A multi-label binary cross-entropy function is employed for model training, defined as,  
\begin{equation}
\label{Eq:supervisedlabelloss}
\begin{split}
&\mathcal{L}(Y,M(X))\\&=-\frac{1}{BC}\sum^{B}_{i=1}\sum^{C}_{c=1}(1-y_{i,c})\log(1-p_{i,c})+y_{i,c}\log p_{i,c},
\end{split}
\end{equation}
where $X=\{x_i\}_{i=1}^{B},x_i \in \mathbb{R}^{12 \times L}$ are the ECG recordings in the current mini-batch, $L$ is the signal length and $Y=\{y_i\}_{i=1}^{B}$ is the corresponding ground truths. $p_{i,c}$ is the model prediction on class $c$ and $C$ is the number of categories. During model training, a held-out validation set is used for early-stop model validation. The best-performing model on the validation set is used for downstream tasks on small-scale datasets.

\subsection{Random-Deactivation Low-Rank Adaptation}
Recent studies have demonstrated that low-rank adaptation (LoRA) can drastically decrease computation and storage costs in large-scale neural network fine-tuning while achieving promising performance on downstream tasks\cite{hu2022lora,zhang2023adaptive,ding2023parameter}. The LoRA method models the incremental update of the pre-trained weights by the matrix multiplication of two low-rank matrices. For a hidden layer output $h=WX$, the LoRA forward process is defined as,
\begin{equation}
\label{Eq:LoRA}
h = (W_0 + \triangle W)X = (W_0 + BA)X, 
\end{equation}
where $W_0, \triangle W \in\mathbb{R}^{d_1 \times d_2}$, $B \in\mathbb{R}^{d_1 \times r}$ and $A \in\mathbb{R}^{r \times d_2}$, and the rank $r\ll \min(d_1,d_2)$. The LoRA freezes the pre-trained weight $W_0$ during model training and only optimizes the low-rank matrices $A$ and $B$, which greatly reduces the number of trainable parameters during model training\cite{hu2022lora}. However, the incremental updates of low-rank matrices are inadequate for achieving optimal performance on downstream datasets\cite{zi2023delta,zhang2023increlora}. To bridge the performance gap efficiently, we propose a novel random-deactivation low-rank adaptation (RD-LoRA) method, which randomly activates or deactivates the low-rank matrices in each trainable layer with a given probability $p$. To be specific, the forward process of the proposed RD-LoRA can be defined as,
\begin{equation}
\label{Eq:RD_LoRA}
h=(W_0 + \delta BA)X, \delta=
\begin{cases}
1, & z\geq p\\
0,& z < p
\end{cases},
\end{equation}
where $\delta \sim B(\delta, 1-p)$ can be regarded as a binary gate controlled by a random variable $z$ following a uniform distribution $U(0,1)$. In the training stage, the multi-label binary cross-entropy loss defined in Eq.\ref{Eq:supervisedlabelloss} is employed for parameter optimization. In the testing stage, for input data $X_t$ and the pre-trained weight $W_0$, the expectation of the output of the RD-LoRA can be given as, 
\begin{equation}
\label{Eq:expectation}
\mathbb{E}_{\delta \sim B(\delta, 1-p) }\left[h_t\right] = \mathbb{E}_{\delta \sim B(\delta, 1-p)}\left[(W_0 + \delta BA)X_t\right].
\end{equation}
 Considering that the test data $X_t$ is independent from the network parameters, the covariance matrix $Cov(X_t, W_0 + \delta AB)$ equals zero, and thus,
\begin{equation}
\label{Eq:expectation_2}
\begin{split}
\mathbb{E}_{\delta \sim B(\delta, 1-p)}\left[h_t\right] &= \mathbb{E}_{\delta \sim B(\delta, 1-p)}\left[(W_0 + \delta BA)X_t\right]\\&=(\mathbb{E}\left[W_0\right]+(1-p)\mathbb{E}\left[BA\right])\mathbb{E}\left[X_t\right]\\&=(W_0+(1-p)BA)X_t.
\end{split}
\end{equation}
Similar to LoRA, the low-rank matrices are merged into the pre-trained weight $W_0$ in the testing stage to avoid extra inference costs, and the random-drop operation is deactivated. According to Eq.\eqref{Eq:expectation_2}, to ensure the expected output will be the same as the output with RD-LoRA, the merged matrix should be computed as, 
\begin{equation}
\label{Eq:merge}
W = W_0 + (1-p)BA.
\end{equation}
After merging the low-rank matrices into the pre-trained weights of different layers, the final network can be viewed as an ensemble of all possible sub-networks during model training, which improves its stability and generalization performance on unseen test datasets. Additionally, randomly deactivating some low-rank matrices avoids the computation of update matrices in some layers, which improves training speed in the low-rank adaptation of large-scale models.
\subsection{Ensemble Optimization Properties of the RD-LoRA}
In this section, we briefly analyze the ensemble properties of the proposed RD-LoRA. Here, we simply consider a network $M$ with $n$ fully-connected layers, defined as $M(X)=\prod_{i=1}^{n} W^{i}_0 X$, where $X$ is the input data and $W^{i}_0\in\mathbb{R}^{c_{out}\times c_{in}}$ is the pre-trained weight matrix at the $i$-th layer. During model training, a convex loss function $\mathcal{L}(Y,M(X))$ is employed for parameter optimization. When the RD-LoRA is activated, the expectation of the loss function $\mathbb{E}_{\delta\sim B(\delta, 1-p)}\left[\mathcal{L}(Y,M(X))\right]$ can be given as,
\begin{equation}
\label{Eq:finalloss}
\begin{split}
&\mathbb{E}_{\delta\sim B(\delta, 1-p)}\left[\mathcal{L}(Y,M(X))\right]\\&= 
(1-p)^n\mathcal{L}(Y,\prod_{i=1}^{n}(W_0^{i}+B^iA^i)X)\\&+\sum_{j=1}^{n}\left[p(1-p)^{n-1}\mathcal{L}(Y,\prod_{i=1,i\neq j}^{n}(W_0^{i}+B^iA^i)W_0^{j})X\right]\\&+\cdots + p^n\mathcal{L}(Y,\prod_{i=1}^{n} W^{i}_0X),
\end{split}
\end{equation}
where the low-rank matrices $\{A^i\}_{i=1}^n$ and $\{B^i\}_{i=1}^n$ are trainable while the pre-trained weights $\{W^i_0\}_{i=1}^n$ are frozen. Eq.\eqref{Eq:finalloss} can be regarded as a weighted mean of the losses of different sub-networks, which are minimized during model training. The number of activated low-rank matrices of the sub-networks is lower than the entire network. Consequently, the training costs of the sub-networks are lower than those of the entire network. In the testing stage, all the low-rank matrices are merged into the pre-trained weights, which generates an ensemble model combing all the possible sub-networks. The testing loss can be estimated as 
\begin{equation}
\label{Eq:test_expect}
\begin{split}
&\mathcal{L}(Y,\mathbb{E}_{\delta \sim B(\delta, 1-p)}\left[M(X_t)\right])\\&=\mathcal{L}\left (Y,\mathbb{E}_{\delta \sim B(\delta,1-p)}\left[\prod_{i=1}^{n}(W^{i}_0+\delta^i B^iA^i)X_t\right] \right )\\&=\mathcal{L}(Y,\prod_{i=1}^{n}(W^{i}_0+(1-p)B^iA^i)X_t).
\end{split}
\end{equation}
In this paper, the multi-label binary cross-entropy loss with sigmoid activation $\sigma(M(X))=\left[\sigma(M(X))_1,\sigma(M(X))_2,\cdots \sigma(M(X))_C\right]$ is convex according to the second-order condition of convexity, where $C$ is the number of categories. Specifically, the Hessian matrix of $\mathcal{L}(Y,\sigma(M(X)))$ is diagonal and the $c$-th element of the main diagonal can be given as,
\begin{equation}
\label{Eq:convex}
\frac{\partial^2 \mathcal{L}(Y,\sigma(M(X)))}{\partial M(X)_c^2}=\sigma(M(X))_c(1-\sigma(M(X))_c)\geq 0,
\end{equation}
where $Y=\left[y_1,y_2,\cdots y_C\right]$, $y_c\in\{0,1\}$ and $\sigma(M(X))=(1+e^{-M(X)})^{-1}$. According to Eq \eqref{Eq:convex}, the Hessian matrix of $\mathcal{L}(Y,\sigma(M(X)))$ is positive semidefinite, demonstrating the convexity of the loss function. Based on Jensen's inequality, the loss of any
ensemble average is smaller than the average loss of the ensemble components,
\begin{equation}
\label{Eq:jensen}
\mathcal{L}(Y,\mathbb{E}_{\delta \sim B(\delta, 1-p)}\left[M(X_t)\right]) \leq \mathbb{E}_{\delta \sim B(\delta, 1-p)}\left[\mathcal{L}(Y,M(X_t))\right].
\end{equation}
In the training stage, the proposed RD-LoRA optimizes the parameters of multiple sub-networks and generates an ensemble network in the testing stage, improving the model performance on the testing data.
\subsection{Efficient One-Shot Rank Allocation}
Another limitation of LoRA is that it prespecifies the same rank for all low-rank incremental matrices, neglecting that their importance in model training varies across layers. In response to this limitation, AdaLoRA\cite{zhang2023increlora} and IncreLoRA\cite{zhang2023adaptive} proposed to dynamically adjust the ranks of different incremental matrices during model training based on their importance, which improved the low-rank adaptation performance. However, these dynamic methods require continuous calculation of the importance of all low-rank matrices in each iteration, significantly increasing the computation time. Additionally, their rank allocation processes are based on the singular value decomposition (SVD) theory and thus require an extra regularization loss to force the orthogonality of the low-rank matrices. This property introduces extra hyper-parameters and computation costs. Here, we propose an efficient one-shot rank allocation method to overcome the computation inefficiency of the existing dynamic methods.

Above all, we introduce some preliminaries about how to estimate the importance of weights in neural networks. Based on the first-order Taylor expansion, the importance of a weight matrix can be computed by the error induced by removing it from the network\cite{molchanov2019taylor}, defined as,
\begin{equation}
\label{Eq:importance}
\begin{split}
I(W^i)&=\frac{1}{N_e}\sum_{j=1}^{N_e}(\mathcal{L}(Y,M(X))-\mathcal{L}_{W^{i}(j)=0}(Y,M(X)))^2 \\&\approx \left\|\frac{\partial \mathcal{L}(Y,M(X))}{\partial W^i} \odot W^i\right\|_2^2,
\end{split}
\end{equation}
where $W^i(j)$ is the $j$-th element in the weight matrix $W^i$, $N_e$ is the number of elements in $W^i$ and $\odot$ is the Hadamard product. However, the gradient matrix $\frac{\partial \mathcal{L}(Y,M(X))}{\partial W^i}$ can not be obtained because $W^i$ is frozen during the low-rank training process. To solve this problem, we approximate it using its incremental update $\triangle W^i$, which can be computed by low-rank matrices $A^i$ and $B^i$ using the Eq.\eqref{Eq:LoRA}. 
\begin{equation}
\label{Eq:estimate_W}
\begin{split}
&\frac{\partial \mathcal{L}(Y,M(X))}{\partial W^i}\propto-\frac{1}{\eta}\triangle W^i =\frac{1}{\eta}(B_{t}^{i} A_{t}^{i}-B_{t+1}^{i} A_{t+1}^{i})\\
&= \frac{1}{\eta}\left[B_{t}^{i} A_{t}^{i}-(B_t^{i}-\eta \frac{\partial \mathcal{L}(Y,M(X))}{\partial B_{t}^{i}})(A_t^{i}-\eta \frac{\partial \mathcal{L}(Y,M(X))}{\partial A_t^{i}})\right]\\&=B_t^{i}\frac{\partial \mathcal{L}(Y,M(X))}{\partial A_t^{i}}+\frac{\partial \mathcal{L}(Y,M(X))}{\partial B_t^{i}}A_n^{i} \\&- \eta\frac{\partial \mathcal{L}(Y,M(X))}{\partial B_t^{i}}\frac{\partial \mathcal{L}(Y,M(X))}{\partial A_t^{i}},
\end{split}
\end{equation}
where $A^i_{t}$ and $B^i_{t}$ are the low-rank matrices at training round $t$, constant $\eta$ is the learning rate and $W_0$ is the pre-trained weight. Although Eq.\eqref{Eq:estimate_W} enables importance score estimation during model training, iterative matrix multiplication induces a heavy computation burden. Hence, we propose to simplify the estimation function Eq.\eqref{Eq:estimate_W} and compute the importance score in a 'one-shot' manner. Specifically, we only use the first gradient-backpropagation process to achieve the entire rank allocation process and fix the ranks of different low-rank matrices during the remaining training iterations. In the first backpropagation process, the low-rank matrices $\{A^i\}^{n}_{i=1}$ are initialized from a normal distribution $N(0,\sigma^2)$ and $\{B^i\}^{n}_{i=1}$ are initialized to zero. Consequently, the gradient of $\{A^i\}^{n}_{i=1}$ at the $0$-th (first) iteration is zero according to Eq.\eqref{Eq:LoRA}. Based on the above initialization conditions, Eq.\eqref{Eq:estimate_W} at the $0$-th iteration can be rewritten as,
\begin{equation}
\label{Eq:update_W_simple}
\begin{split}
\frac{\partial \mathcal{L}(Y,M(X))}{\partial W_0^i}&\propto-\frac{1}{\eta}\triangle W_0^i = \frac{\partial \mathcal{L}(Y,M(X))}{\partial B_0^{i}}A_0^{i}, \\&\frac{\partial \mathcal{L}(Y,M(X))}{\partial A_0^{i}}=0, B_0^i = 0,
\end{split}
\end{equation}
where $\{W_0^i\}_{i=1}^{n}$ are the pre-trained weight matrices in the backbone model $M(X)$. Then, the importance score of the pre-trained weight $W^i_0$ can be approximated as, 
\begin{equation}
\label{Eq:importance_final}
I(W_0^i) \approx \hat{I}(W_0^i) = \left\|\left (\frac{\partial \mathcal{L}(Y,M(X))}{\partial B_0^{i}}A_0^{i} \right )\odot W^{i}_0\right\|_2^2.
\end{equation}
Then, we sort the importance $\hat{I}(W_0^i)$ of all pre-trained matrices in descending order and allocate different ranks for their low-rank matrices. Here, we assume the ranks of the incremental matrices corresponding to the important weights should be higher than those of the incremental matrices associated with the unimportant weights. The allocated rank $r^i$ of the incremental matrices of the pre-trained weight $W^i_0$ is defined as,
\begin{equation}
\label{Eq:rank}
r^{i}=
\begin{cases}
r, & \text{$\hat{I}(W_0^i)$ in the top-$k$ of $\{\hat{I}(W_0^i)\}_{i=1}^n$}\\
\frac{1}{2}r,& \text{otherwise}
\end{cases}, k=nc,
\end{equation}
where $r$ is an initial rank, and $c$ is a hyper-parameter that controls the number of important weight matrices. 
Note that the allocated ranks $\{r^{i}\}_{i=1}^{n}$ are fixed during the remaining iterations, and the low-rank matrices ($\{B^i\}^{n}_{i=1}$, $\{A^i\}^{n}_{i=1}$) are reset based on their allocated ranks. Eq.\eqref{Eq:importance_final} is only computed at the 0-th iteration, which avoids numerous matrix multiplication. In addition, the proposed rank allocation process does not require constraint on the orthogonality of low-rank matrices. In summary, the above advantages allow the proposed method to have a faster training speed compared to existing dynamic methods, such as AdaLoRA\cite{zhang2023increlora} and IncreLoRA\cite{zhang2023adaptive}. 

\subsection{Lightweight Semi-Supervised Learning}
Semi-supervised learning (SSL) is an efficient tool for model performance enhancement when large-scale unlabeled data is available\cite{semi2006,berthelot2019mixmatch}. Recently, many studies utilized label guessing and consistency regularization to further improve the model performance in SSL tasks, such as FixMatch\cite{sohn2020fixmatch}, FlexMatch\cite{zhang2021flexmatch} and SoftMatch\cite{chen2023softmatch}. However, the above two techniques require the output predictions of the weak and strong-augmented unlabeled samples, which induces extra computation costs. Consequently, traditional SSL methods usually exhibit much higher memory costs and longer training time than naive supervised models. Here, we introduce a lightweight but effective SSL method without extensive consistency training and pseudo-label guessing. 

The main drawback of fully-supervised learning on small datasets is over-fitting, while semi-supervised learning alleviates this problem by utilizing large-scale unlabeled data. However, the heavy computation burden limits their application in real-world scenarios. Consequently, a natural question is: How can we alleviate the over-fitting problem in a lightweight but effective manner? Our solution is to update the batch normalization (BN) layers in a semi-supervised manner using both labeled and unlabeled data. Subsequently, the unlabeled data is released, and only the labeled data is forwarded to the self-attention and classification blocks for loss computation. \textcolor{black}{For labeled inputs $\{x_{b}^{i}\}_{i=1}^{N_B}$ and unlabeled inputs $\{x_{u}^{i}\}_{i=1}^{N_U}$, the mean value $\mu$ and the variance $\sigma$ of the semi-supervised BN layers in the convolution blocks can be updated as,}
\begin{equation}
\label{Eq:semi_bn_mean}
\mu =\frac{\gamma}{N_B}\sum_{i=1}^{N_B} x_b^i + \frac{1-\gamma}{N_U}\sum_{i=1}^{N_U} x_u^i,
\end{equation}
\begin{equation}
\label{Eq:semi_bn_sigma}
\sigma =\frac{\gamma}{N_B}\sum_{i=1}^{N_B}(x^i_b-\mu)^2 + \frac{1-\gamma}{N_U}\sum_{i=1}^{N_U}(x^i_u-\mu)^2,
\end{equation}
\textcolor{black}{where $N_B$ and $N_U$ are the numbers of labeled and unlabeled samples in the current mini-batch, and $\gamma=\frac{N_B}{N_B+N_U}$. Note that $N_B$ equals $N_U$ in this study, thus $\gamma=0.5$. With only limited labeled data $x_b$, the estimated mean $\mu_B=\frac{1}{N_B}\sum_{i=1}^{N_B} x_b^i$ and variance $\sigma_B=\frac{1}{N_B}\sum_{i=1}^{N_B} (x_b^i-\mu_B)^2$ in traditional BN are prone to be influenced by the over-fitting problem according to the law of large numbers.  On the contrary,  the proposed semi-supervised BN alleviates the problem by utilizing large-scale unlabeled data $x_u$ for parameter estimation, which improves the model performance on unseen distributions.} Since the BN layers do not exist in the self-attention and classification blocks, we only forward the labeled features to them to reduce memory cost and training time. Compared with the SOTA methods in semi-supervised learning, the proposed CE-SSL discards the label guessing and the consistency regularization modules. However, the results demonstrate that it achieves similar or even better performance on four datasets. More importantly, its computation costs are much lower than those of the SOTA methods, including less memory consumption and faster training speed.
\begin{algorithm*}[t]
  \caption{CE-SSL algorithm}
  \label{alg::conjugateGradient}
  \begin{algorithmic}[1]
    \Require
\renewcommand{\algorithmicrequire}{\textbf{}}
    \Require - Labeled dataset $D_B=\{X_b,Y_b\}$ and unlabeled dataset $D_U=\{X_u\}$;
    \Require - Pre-trained model $M_0 = \{W_{0}^{i}\}_{i=1}^{n}$; Initial rank $r$; The ratio of important weights $c$; The random-deactivation probability $p$; Batch sizes of the labeled samples ($N_B=64$) and the unlabeled samples ($N_U=64$).
    \Ensure Adapted model $M$ with the updated parameters $\{W^{i} = W_0^{i} + (1-p)A^iB^i\}_{i=1}^{n}$; 
    \State \textcolor{gray}{One-shot rank allocation}
    \State Compute the importance of each pre-trained weight using the Eq.\eqref{Eq:importance_final} and the labeled dataset $D_B$;
    \State Based on the initial rank $r$ and the ratio $c$, allocate the final rank $r^i$ of the incremental matrices ($A^i$,$B^i$) of the pre-trained weight $W^i_0$ using Eq.\eqref{Eq:rank}.
        \For {1 to $iteration$} 
          \State sample labeled data $\{x_b,y_b\}$ from $D_B$; 
          \State sample unlabeled data $\{x_u\}$ from $D_U$;
          \State apply data augmentation to $x_b$ and $x_u$;
          \State \textcolor{gray}{Lightweight semi-supervised learning}
          \State Based on Eq.\eqref{Eq:semi_bn_mean} and Eq.\eqref{Eq:semi_bn_sigma}, update the semi-supervised batch-normalization layers in the convolution blocks using the labeled data $x_b$ and the unlabeled data $x_u$.
          \State release the unlabeled data $x_u$ in the GPU memory
          \State \textcolor{gray}{Random-deactivation low-rank adaptation}
          \State initialize $h_0=x_b$
           \For {$i = 1,2,...n$} 
           \State sample $\delta_i$ from the Bernoulli distribution $B(\delta, 1-p)$
           \State  $h_i = (W_0^{i} + \delta B^iA^i)h_{i-1}$
          \EndFor     
          \State Based on the model output $h_n$ and the ground-truth $y_b$, compute the supervised multi-label binary cross-entropy loss using Eq.\eqref{Eq:supervisedlabelloss}.
          \State apply an early-stop strategy to avoid overfitting;
    \EndFor
    \State Merge the incremental matrices into the pre-trained weights, as $\{W^{i} = W_0^{i} + (1-p)B^iA^i\}_{i=1}^{n}$;
\end{algorithmic}
\end{algorithm*}

\subsection{Signal Pre-Processing and Data Augmentation}
\label{sec:preprocess}
Artifact removal and data augmentation are two factors that play important roles in model performance. Firstly, we introduce the signal pre-processing pipeline employed in the proposed framework. The ECG recordings from the CODE-15\% and CODE-full databases are first resampled to a 400Hz sampling rate following the configuration of the dataset provider\cite{ribeiro2020automatic}. The sampling rate of the recordings from the four downstream databases remains unchanged. Firstly, the length of all recordings is normalized into 6144 samples by zero-padding. Subsequently, a band-pass filter (1-47Hz) is applied to remove the power-line interference and baseline drift. Then, the pre-processed signals are normalized using z-score normalization. Secondly,  CutMix\cite{yun2019cutmix} is employed for labeled data augmentation. Since the sample generation process of CutMix requires true labels that are absent in the unlabeled data, we employed the ECGAugment\cite{zhoupami2023} for unlabeled data augmentation, which generates new samples by randomly selecting a transformation to perturb the pre-processed signals. Note that only the weak-augmentation module in the ECGAugment is employed.

\section{Experiments and Datasets}
\label{sec:experiment}
In this section, we utilized two large-scale datasets to pre-train the backbone and evaluated the performance of our CE-SSL on four downstream datasets. First, an openly available ECG dataset (CODE-15\%) collected by the Telehealth Center of the Universidade Federal de Minas Gerais\cite{ribeiro2019tele,ribeiro2020automatic} is used for the base backbone pre-training. The CODE-15\% dataset contains 15\% of the ECG data from the restricted CODE-full dataset\cite{ribeiro2020automatic,Leilu2024}, which are used for pre-training the medium and large backbones. Specifically, there are 345779 ECG recordings from 233770 patients in the CODE-15\% dataset, alongside six CVD labels: 1st degree AV block (1dAVb), left bundle branch block (LBBB), right bundle branch block (RBBB), sinus tachycardia (ST), atrial fibrillation (AF) and sinus bradycardia (SB). Each ECG recording lasts 7-10s, and the sampling rate is 300-600 Hz. Subsequently, we use four small datasets for downstream model retraining and evaluation: the Georgia 12-lead ECG Challenge (G12EC) database\cite{alday2020classification}, the Chapman-Shaoxing database\cite{zheng202012}, the Ningbo database\cite{zheng2020optimal}, and the Physikalisch-Technische Bundesanstalt (PTB-XL) database\cite{wagner2020ptb}. Specifically, the G12EC database contains 10344 ECG recordings from 10,344 people, and the PTB-XL database comprises 21837 recordings from 18885 patients. The Chapman database contains 10,646 recordings from 10646 patients, and the Ningbo database encompasses 40258 recordings from 40258 patients. Only 34,905 recordings in the Ningbo database are publicly available\cite {alday2020classification}. The recordings from the four downstream databases are around 10 seconds, and the sampling rate is 500 Hz. Additionally, each database contains over 17 different CVDs, and multiple CVDs can be identified from one ECG segment simultaneously.

The base backbone model is first pre-trained on the CODE-15\% dataset using the ECG recordings and the corresponding multi-label ground truths. The architecture
of the backbone is shown in Fig.\ref{fig:flowchart}. AdawW optimizer\cite{loshchilov2017decoupled} is used for the pre-training process with a learning rate of 1e-3 and a batch size of 1024. Then, the pre-trained backbone is retrained on the four downstream datasets using different methods under limited supervision. Taking the G12EC database as an example, the ECG recordings are split into a training set and a held-out test set in a ratio of 0.9: 0.1. Then, the training set is divided into a labeled training set and an unlabeled training set in a ratio of 0.05: 0.95. A validation set is randomly sampled from the labeled training set and accounts for 20\% of it, which is used for selecting the best-performing model during training.  For model comparisons, we reproduce several baseline models in semi-supervised learning: ReMixMatch\cite{berthelot2019remixmatch}, FixMatch\cite{sohn2020fixmatch}, FlexMatch\cite{zhang2021flexmatch}, SoftMatch\cite{chen2023softmatch}, MixedTeacher\cite{zhang2022semi}, Adsh\cite{guo2022class}, SAW\cite{lai2022smoothed}. Additionally, we integrate the state-of-the-art parameter-efficient methods (LoRA\cite{hu2022lora}, DyLoRA\cite{2023dylora}, AdaLoRA\cite{zhang2023adaptive}, IncreLoRA\cite{zhang2023increlora}) with FixMatch for comparisons.

We comprehensively evaluate the model performance of various methods using multiple metrics and training costs. Since multiple CVDs can be detected from one recording simultaneously, we used metrics on multi-label classification. Our metrics include ranking loss, coverage, mean average precision (MAP), macro AUC, macro $G_{\beta=2}$ score, and macro $F_{\beta=2}$ score. We set the $\beta$ value to be 2 for all the corresponding experiments following the configurations provided in ref.\cite{strodthoff2020deep}. Lower values indicate better model performance in the first two metrics (ranking loss and coverage), whereas the inverse holds for the remaining metrics. A detailed computation process for the metrics can be found in Supplementary Materials Section 1. Additionally, we report the training costs of different models. Specifically, the peak GPU memory footprint during model training (Mem), the number of trainable parameters (Params), and the average training time for each optimization iteration (Time/iter) are presented. The higher the number of trainable parameters, the higher the parameter storage consumption. Note that the number of trainable parameters of CE-SSL can be adjusted by the initial rank $r$. Lower ranks indicate fewer trainable parameters. The AdamW optimizer\cite{loshchilov2017decoupled} is used under a learning rate of 1e-3 and a batch size of 64 for all the compared methods. All the experiments are conducted in a single NVIDIA A6000 graphics processing unit using the Pytorch library. 

\section{Results and Discussion}
\label{sec:discussion} 
\subsection{Analysis of the CVDs Detection Results}
Table \ref{tab:compare_semi} shows that our proposed CE-SSL achieved superior detection performance on four datasets with the lowest computational costs compared with the baseline models. For example, in the G12EC dataset, CE-SSL with $r=16$ achieves a macro $F_{\beta=2}$ of 0.551$\pm$0.017, which is 4.1\% larger than the second-best model's (FixMatch) performance. In Supplementary Materials Section 2, we present the detection performance of different models on each CVD. The results demonstrate that CE-SSL ranks the best in some CVDs, such as atrial fibrillation and first-degree AV block. It also achieves comparable performance to the compared methods in the remaining CVDs. Regarding computational costs, it requires 33.5\% less training time than MixedTeacher, occupies 29.3\% less GPU memory than Adsh, and has only 5.8\% of the trainable parameters found in them. When the initial rank $r$ decreases to 4, CE-SSL shows a slight performance drop in four datasets, but the number of trainable parameters further decreases to 1.8\% of the baseline models. This observation indicates the stability and robustness of the CE-SSL under extremely low parameter budgets.

\begin{figure*}[t]
\begin{center}
\includegraphics[width=1\textwidth]{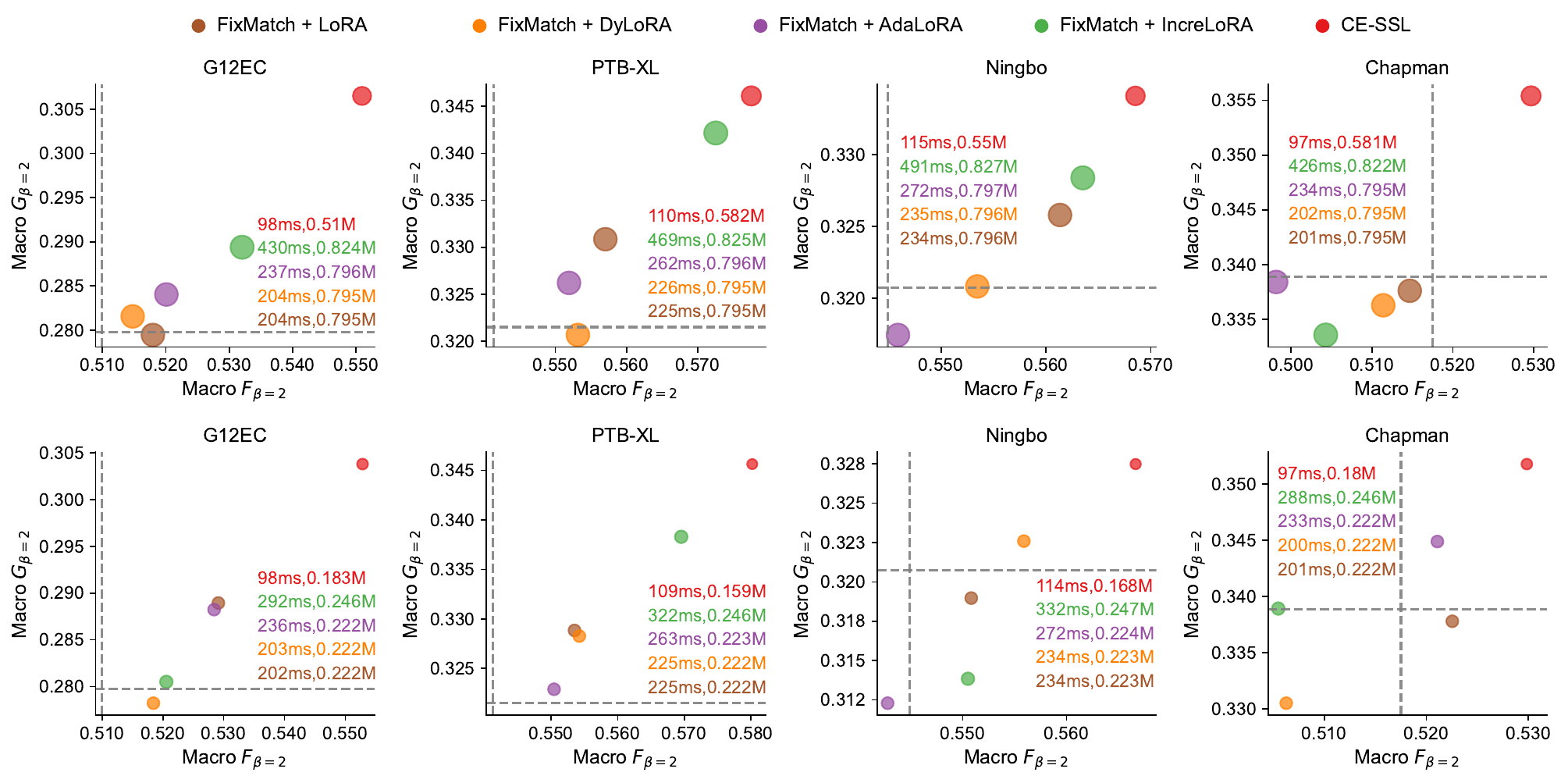}
\end{center}
\caption{Comparison between CE-SSL and parameter-efficient semi-supervised methods on the base backbone. Circles with various colors denote different models, and their size represents the number of trainable parameters. The training time for each optimization iteration (Time/iter) of different methods is also reported. The gray dotted lines represent the performance of the FixMatch baseline without parameter-efficient training (approximately 9.505M trainable parameters). The first row of the figure presents the performance of different models with sufficient parameter budgets ($r=16$), while the second row reports their performance under limited parameter budgets ($r=4$).}
\label{fig:bubble}
\end{figure*}

We further compare the proposed CE-SSL with the parameter-efficient methods, which are integrated with FixMatch for parameter-efficient semi-supervised learning. For example, FixMatch with low-rank adaptation (LoRA) is denoted as 'FixMatch+LoRA'. Similar to the CE-SSL, their budgets for the number of trainable parameters are controlled by the initial rank $r$. As illustrated in Fig. \ref{fig:bubble}, we report their macro-$F_{\beta=2}$ scores, macro-$G_{\beta=2}$ scores, and Time/iter on four datasets at sufficient ($r=16$) and limited ($r=4$) budget levels. The macro-$F_{\beta=2}$ and macro-$G_{\beta=2}$ scores of the FixMatch without parameter-efficient training (Params: 9.505M) are denoted as gray dotted lines. The experiment results indicate that CE-SSL consistently outperforms the other methods on four datasets at different budget levels. Under a sufficient parameter budget ($r=16$), CE-SSL achieves a macro-$G_{\beta=2}$ score of 0.307$\pm$0.016 on the G12EC dataset, which is 2.8\% higher than the FixMatch with LoRA.  When the parameter budget is limited ($r=4$), CE-SSL still outperforms it by 1.5\%. Additionally, CE-SSL achieves the highest training speed and the best performance with the least trainable parameters compared to other parameter-efficient semi-supervised learning frameworks. On the four datasets, CE-SSL requires 50\% less training time compared to other methods. This phenomenon demonstrates the computational efficiency of the proposed CE-SSL. In Supplementary Materials Table S8, we present detailed comparison results on more evaluation metrics, which provide supplementary evidence on the efficiency of the proposed CE-SSL in CVDs detection. Paired t-tests are conducted to evaluate the significance levels of the performance difference between CE-SSL and the aforementioned SOTA methods (Supplementary Materials Section 3 Fig. S1). Based on the calculated two-sided $p$-value, it can be observed that CE-SSL outperforms the baselines at a 0.05 significance level in most datasets and evaluation metrics, which indicates a significant superiority for the proposed CE-SSL framework.

In summary, our experiment results demonstrate the robustness and computational efficiency of the CE-SSL in semi-supervised cardiovascular disease detection at different parameter budget levels. In other words, CE-SSL can enhance the detection performance of ECG-based CVDs detection models without introducing heavy computation burdens.

\begin{table*}[t]
\fontsize{8}{10}\selectfont
\setlength{\tabcolsep}{0.2em}
\begin{center}
\caption{Performance comparisons of CE-SSL and semi-supervised baselines on the base backbone. The average performance on all CVDs within each dataset is shown across six seeds. The standard deviation is also reported for the evaluation metrics.   }
\label{tab:compare_semi}
\scalebox{1}{
\color{black}
\begin{tabular*}{\hsize}{@{}@{\extracolsep{\fill}}lccccccccc@{}}
\toprule
Methods & Params $\downarrow$ & Mem $\downarrow$ & Time/iter $\downarrow$ & Ranking Loss $\downarrow$ & Coverage $\downarrow$ & Macro AUC $\uparrow$ & MAP $\uparrow$ & Macro $G_{\beta=2}$ $\uparrow$ & Macro $F_{\beta=2}$ $\uparrow$\\
\midrule
\multicolumn{10}{c}{\textbf{G12EC Dataset}}\\
\midrule
ReMixMatch &9.505 M&7.699 GB&271 ms&0.178$\pm$0.033&5.635$\pm$0.529&0.759$\pm$0.015&0.310$\pm$0.019&0.184$\pm$0.017&0.387$\pm$0.031\\
MixedTeacher & 9.505 M&3.941 GB&147 ms&0.107$\pm$0.009&4.224$\pm$0.236&0.835$\pm$0.010&0.464$\pm$0.003&0.275$\pm$0.016&0.507$\pm$0.025\\
FixMatch &9.505 M&5.784 GB&187 ms&0.107$\pm$0.006&4.292$\pm$0.163&0.829$\pm$0.004&0.468$\pm$0.009&0.280$\pm$0.010&0.510$\pm$0.016\\
FlexMatch &9.505 M&5.784 GB&187 ms&0.113$\pm$0.005&4.365$\pm$0.133&0.829$\pm$0.009&0.450$\pm$0.022&0.274$\pm$0.019&0.497$\pm$0.035\\
SoftMatch & 9.505 M&5.784 GB&187 ms&0.110$\pm$0.006&4.313$\pm$0.128&0.834$\pm$0.004&0.457$\pm$0.010&0.276$\pm$0.017&0.504$\pm$0.021\\
Adsh & 9.505 M&3.887 GB&207 ms&0.111$\pm$0.003&4.387$\pm$0.129&0.827$\pm$0.005&0.458$\pm$0.007&0.268$\pm$0.009&0.489$\pm$0.013\\
SAW & 9.505 M&5.784 GB&188 ms&0.112$\pm$0.003&4.369$\pm$0.105&0.827$\pm$0.005&0.459$\pm$0.017&0.269$\pm$0.018&0.494$\pm$0.024\\
\textbf{CE-SSL}\textsubscript{$r$=16} & \textbf{0.510 M}&\textbf{2.747 GB}&\textbf{98 ms}&\textbf{0.092$\pm$0.002}&\textbf{3.867$\pm$0.088}&\textbf{0.855$\pm$0.005}&\textbf{0.476$\pm$0.006}&\textbf{0.307$\pm$0.016}&\textbf{0.551$\pm$0.017}\\
\textbf{CE-SSL}\textsubscript{$r$=4} & \textbf{0.183 M}&\textbf{2.743 GB}&\textbf{98 ms}&\textbf{0.089$\pm$0.003}&\textbf{3.804$\pm$0.095}&\textbf{0.853$\pm$0.004}&\textbf{0.467$\pm$0.006}&\textbf{0.304$\pm$0.013}&\textbf{0.553$\pm$0.020}\\
\midrule
\multicolumn{10}{c}{\textbf{PTB-XL Dataset}}\\
\midrule
ReMixMatch &9.505 M&7.699 GB&301 ms&0.068$\pm$0.008&3.699$\pm$0.196&0.827$\pm$0.006&0.350$\pm$0.010&0.212$\pm$0.015&0.423$\pm$0.028\\
MixedTeacher & 9.505 M&3.941 GB&164 ms&0.037$\pm$0.003&2.841$\pm$0.095&0.884$\pm$0.008&0.509$\pm$0.008&0.316$\pm$0.007&0.542$\pm$0.014\\
FixMatch &9.505 M&5.784 GB&208 ms&0.038$\pm$0.001&2.905$\pm$0.061&0.882$\pm$0.004&0.510$\pm$0.006&0.322$\pm$0.007&0.541$\pm$0.007\\
FlexMatch &9.505 M&5.784 GB&209 ms&0.039$\pm$0.001&2.937$\pm$0.048&0.887$\pm$0.005&0.505$\pm$0.005&0.316$\pm$0.008&0.536$\pm$0.007\\
SoftMatch & 9.505 M&5.784 GB&209 ms&0.039$\pm$0.003&2.919$\pm$0.097&0.885$\pm$0.006&0.508$\pm$0.007&0.317$\pm$0.009&0.540$\pm$0.011\\
Adsh & 9.505 M&3.887 GB&316 ms&0.038$\pm$0.002&2.879$\pm$0.054&0.886$\pm$0.004&0.511$\pm$0.005&0.322$\pm$0.008&0.543$\pm$0.015\\
SAW & 9.505 M&5.784 GB&208 ms&0.037$\pm$0.003&2.855$\pm$0.093&0.889$\pm$0.005&0.520$\pm$0.007&0.323$\pm$0.019&0.548$\pm$0.017\\
\textbf{CE-SSL}\textsubscript{$r$=16} & \textbf{0.582 M}&\textbf{2.748 GB}&\textbf{110 ms}&\textbf{0.031$\pm$0.000}&\textbf{2.641$\pm$0.020}&\textbf{0.901$\pm$0.003}&\textbf{0.530$\pm$0.005}&\textbf{0.346$\pm$0.006}&\textbf{0.578$\pm$0.006}\\
\textbf{CE-SSL}\textsubscript{$r$=4} & \textbf{0.159 M}&\textbf{2.744 GB}&\textbf{109 ms}&\textbf{0.030$\pm$0.001}&\textbf{2.626$\pm$0.026}&\textbf{0.899$\pm$0.004}&\textbf{0.526$\pm$0.005}&\textbf{0.346$\pm$0.005}&\textbf{0.580$\pm$0.006}\\
\midrule
\multicolumn{10}{c}{\textbf{Ningbo Dataset}}\\
\midrule
ReMixMatch &9.506 M&7.699 GB&312 ms&0.071$\pm$0.010&3.999$\pm$0.231&0.867$\pm$0.008&0.338$\pm$0.024&0.214$\pm$0.016&0.430$\pm$0.020\\
MixedTeacher & 9.506 M&3.941 GB&173 ms&0.035$\pm$0.002&2.982$\pm$0.077&0.925$\pm$0.006&0.496$\pm$0.020&0.324$\pm$0.018&0.549$\pm$0.028\\
FixMatch &9.506 M&5.784 GB&217 ms&0.035$\pm$0.003&3.025$\pm$0.121&0.922$\pm$0.009&0.493$\pm$0.023&0.321$\pm$0.014&0.545$\pm$0.020\\
FlexMatch &9.506 M&5.784 GB&217 ms&0.037$\pm$0.002&3.078$\pm$0.090&0.921$\pm$0.007&0.489$\pm$0.024&0.318$\pm$0.012&0.544$\pm$0.019\\
SoftMatch & 9.506 M&5.784 GB&217 ms&0.035$\pm$0.001&3.018$\pm$0.049&0.923$\pm$0.005&0.496$\pm$0.024&0.321$\pm$0.014&0.552$\pm$0.020\\
Adsh & 9.506 M&3.887 GB&423 ms&0.035$\pm$0.002&3.007$\pm$0.090&0.921$\pm$0.004&0.492$\pm$0.023&0.318$\pm$0.010&0.545$\pm$0.012\\
SAW & 9.506 M&5.784 GB&215 ms&0.037$\pm$0.001&3.064$\pm$0.036&0.924$\pm$0.004&0.492$\pm$0.024&0.314$\pm$0.010&0.536$\pm$0.016\\
\textbf{CE-SSL}\textsubscript{$r$=16} & \textbf{0.550 M}&\textbf{2.748 GB}&\textbf{115 ms}&\textbf{0.030$\pm$0.001}&\textbf{2.805$\pm$0.063}&\textbf{0.928$\pm$0.002}&\textbf{0.505$\pm$0.019}&\textbf{0.334$\pm$0.011}&\textbf{0.569$\pm$0.014}\\
\textbf{CE-SSL}\textsubscript{$r$=4} & \textbf{0.168 M}&\textbf{2.744 GB}&\textbf{114 ms}&\textbf{0.030$\pm$0.001}&\textbf{2.776$\pm$0.028}&\textbf{0.929$\pm$0.001}&\textbf{0.500$\pm$0.017}&\textbf{0.327$\pm$0.010}&\textbf{0.567$\pm$0.011}\\
\midrule
\multicolumn{10}{c}{\textbf{Chapman Dataset}}\\
\midrule
ReMixMatch &9.504 M&7.699 GB&270 ms&0.101$\pm$0.025&3.574$\pm$0.393&0.839$\pm$0.011&0.403$\pm$0.015&0.246$\pm$0.018&0.428$\pm$0.020\\
MixedTeacher & 9.504 M&3.941 GB&148 ms&0.047$\pm$0.002&2.615$\pm$0.068&0.889$\pm$0.012&0.519$\pm$0.018&0.327$\pm$0.019&0.510$\pm$0.024\\
FixMatch &9.504 M&5.784 GB&186 ms&0.046$\pm$0.004&2.626$\pm$0.096&0.897$\pm$0.006&0.520$\pm$0.009&0.339$\pm$0.012&0.518$\pm$0.025\\
FlexMatch &9.504 M&5.784 GB&185 ms&0.047$\pm$0.004&2.659$\pm$0.103&0.895$\pm$0.006&0.518$\pm$0.008&0.325$\pm$0.010&0.495$\pm$0.019\\
SoftMatch & 9.504 M&5.784 GB&187 ms&0.047$\pm$0.004&2.649$\pm$0.079&0.898$\pm$0.006&0.525$\pm$0.012&0.335$\pm$0.011&0.511$\pm$0.021\\
Adsh & 9.504 M&3.887 GB&207 ms&0.046$\pm$0.004&2.621$\pm$0.117&0.896$\pm$0.005&0.528$\pm$0.008&0.335$\pm$0.013&0.517$\pm$0.020\\
SAW & 9.504 M&5.784 GB&185 ms&0.049$\pm$0.003&2.699$\pm$0.072&0.897$\pm$0.007&0.524$\pm$0.009&0.333$\pm$0.012&0.510$\pm$0.020\\
\textbf{CE-SSL}\textsubscript{$r$=16} & \textbf{0.581 M}&\textbf{2.748 GB}&\textbf{97 ms}&\textbf{0.040$\pm$0.002}&\textbf{2.483$\pm$0.055}&\textbf{0.896$\pm$0.006}&\textbf{0.536$\pm$0.004}&\textbf{0.355$\pm$0.005}&\textbf{0.530$\pm$0.008}\\
\textbf{CE-SSL}\textsubscript{$r$=4} & \textbf{0.180 M}&\textbf{2.743 GB}&\textbf{97 ms}&\textbf{0.038$\pm$0.002}&\textbf{2.418$\pm$0.049}&\textbf{0.898$\pm$0.005}&\textbf{0.526$\pm$0.006}&\textbf{0.352$\pm$0.009}&\textbf{0.530$\pm$0.012}\\
\bottomrule
\end{tabular*}
}
\end{center}
\end{table*}

\subsection{Effect of the Ratio of Labeled Samples}
\begin{figure*}[h]
\begin{center}
\includegraphics[width=1\textwidth]{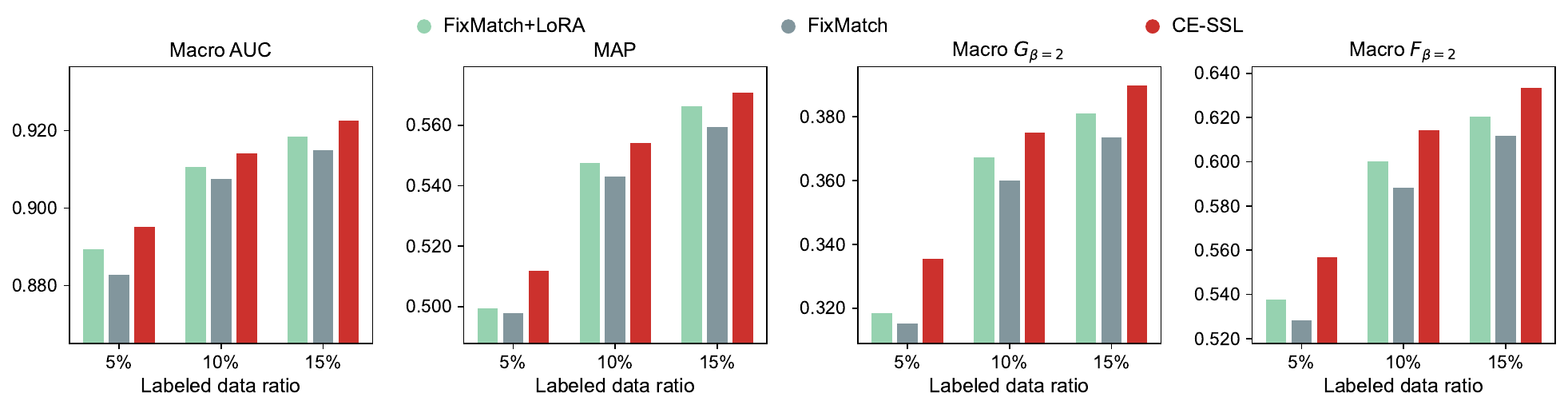}
\end{center}
\caption{Effect of the ratio of labeled samples for model training. We adjust the ratio of the labeled samples in the dataset from 0.05 to 0.15 and report the averaged performance of different models across four datasets and six random seeds.}
\label{fig:sens_ratio}
\end{figure*}

Here, we compare the proposed CE-SSL and baseline models under various ratios of labeled samples in the datasets. Specifically, we adjust the ratio of the labeled samples in the dataset from 5\% to 15\% and present the averaged performance of different models on the four datasets in Fig. \ref{fig:sens_ratio}. The experiment results demonstrate the superiority of the proposed CE-SSL compared with FixMatch and FixMatch with LoRA under various ratios of the labeled data, especially when the ratio is low. As the ratio decreases from 15\% to 5\%, the performance advantage of CE-SSL over other models becomes more significant. When using 15\% labeled data, CE-SSL achieves improvements of 1.3\% on the macro $F_{\beta=2}$ compared to FixMatch with LoRA. In contrast, CE-SSL outperforms it by 1.9\% on the macro $F_{\beta=2}$ using 5\% labeled data.  In Supplementary Materials Fig. S4, we also compare CE-SSL with other baseline models, where CE-SSL consistently outperforms them in CVDs detection under various labeled ratios.

\subsection{Rank Initialization in the One-Shot Rank Allocation} 
\begin{figure*}[h]
\begin{center}
\includegraphics[width=1\textwidth]{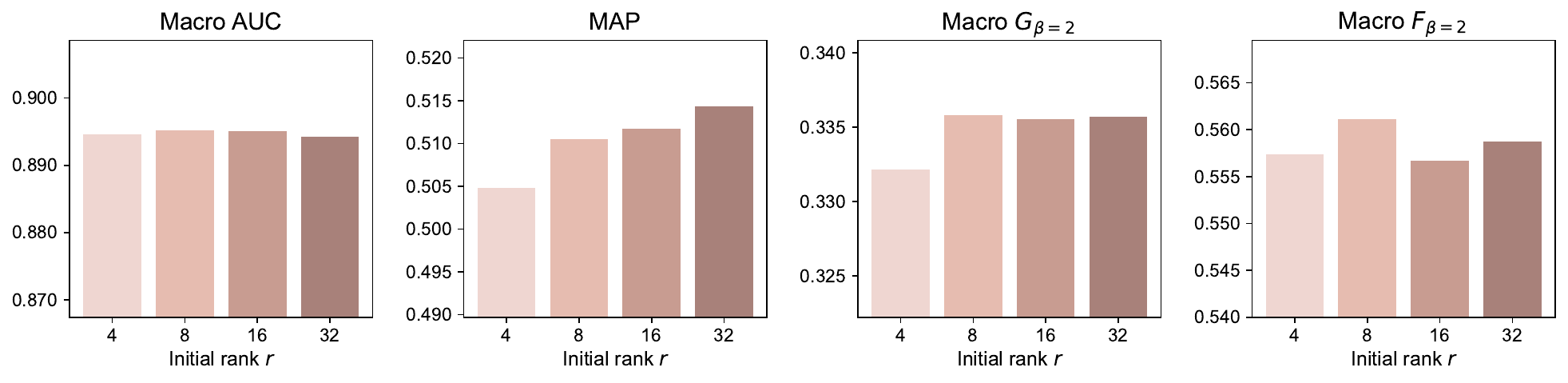}
\end{center}
\caption{Effect of the rank initialization. Averaged performance of the CE-SSL across four datasets and six random seeds under different initial ranks $r$.}
\label{fig:sens_rank}
\end{figure*}

Rank initialization is an important component in low-rank adaptation, which controls the number of trainable parameters during model training. In this section, we adjust the initial rank from 4 to 32 and present the averaged model performance on the four datasets in Fig.\ref{fig:sens_rank}. Note that the labeled ratio is set to 5\%. The results indicate that CE-SSL with high initial ranks ($r=16,32$) achieves better performance than that with low initial ranks ($r=4,8$). This is because the model with higher ranks has more trainable parameters and thus demonstrates a larger capacity during training.

\subsection{Effect of the Number of Important Weight Matrices}
\begin{figure*}[h]
\begin{center}
\includegraphics[width=1\textwidth]{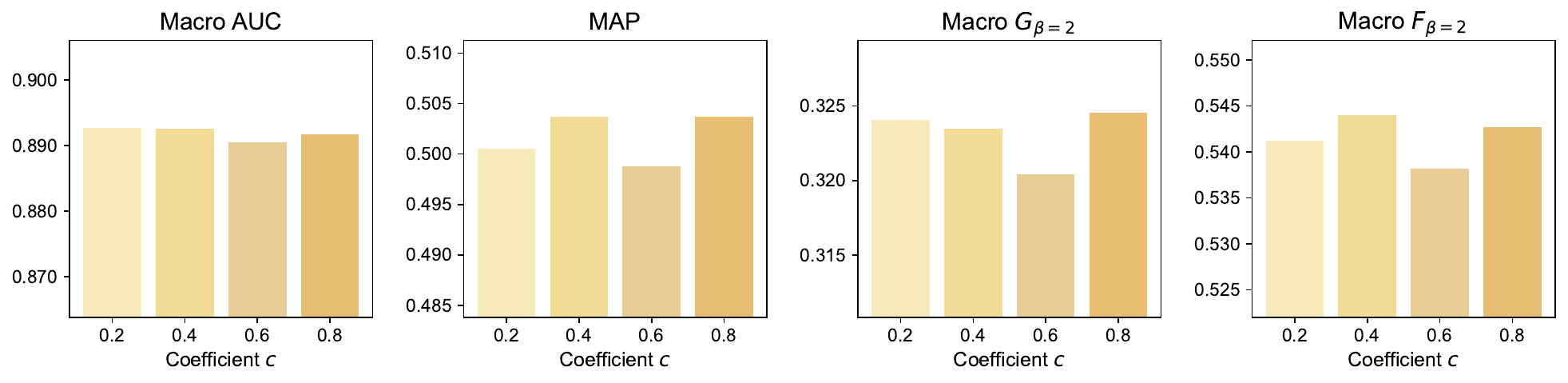}
\end{center}
\caption{Effect of the ratio of important weight matrices. We adjust the ratio of the important weight matrices to the total number of weight matrices and report the averaged performance across four datasets and six random seeds. Important weights are adapted with rank $r$ while the remaining weights are adapted with rank $\frac{1}{2}r$.}
\label{fig:sens_compress}
\end{figure*}

Based on the proposed one-shot rank allocation, CE-SSL allocates a rank $r$ to the incremental matrices with high importance and a rank $r/2$ to the matrices with low importance. The ratio of the important matrices to the total number of pre-trained matrices is defined as the coefficient $c$. The higher the coefficient is, the higher the ratio of the important matrices. In Fig.\ref{fig:sens_compress}, we adjust the coefficient from 0.2 to 0.8 and report the averaged model performance across four datasets. Note that the labeled ratio is set to 5\%, and the initial ranks $r$ for all the low-rank matrices are set to 16. It can be observed that the performance of the proposed model is relatively insensitive to the changes in the $c$. In Supplementary Materials Fig. S5, we visualize the rank distribution generated by the proposed method under various coefficients $c$. When the ratio of important matrices decreases from 0.8 to 0.2, the proposed method allocates more ranks to the self-attention and classification blocks than the convolution blocks. This phenomenon indicates that the deep modules exhibit higher importance than the shallow modules during model training, which aligns with the conclusions made by previous studies\cite{li2021prefix,zhang2023adaptive}. 

\subsection{Effect of the Deactivation Probability}
\begin{figure*}[h]
\begin{center}
\includegraphics[width=1\textwidth]{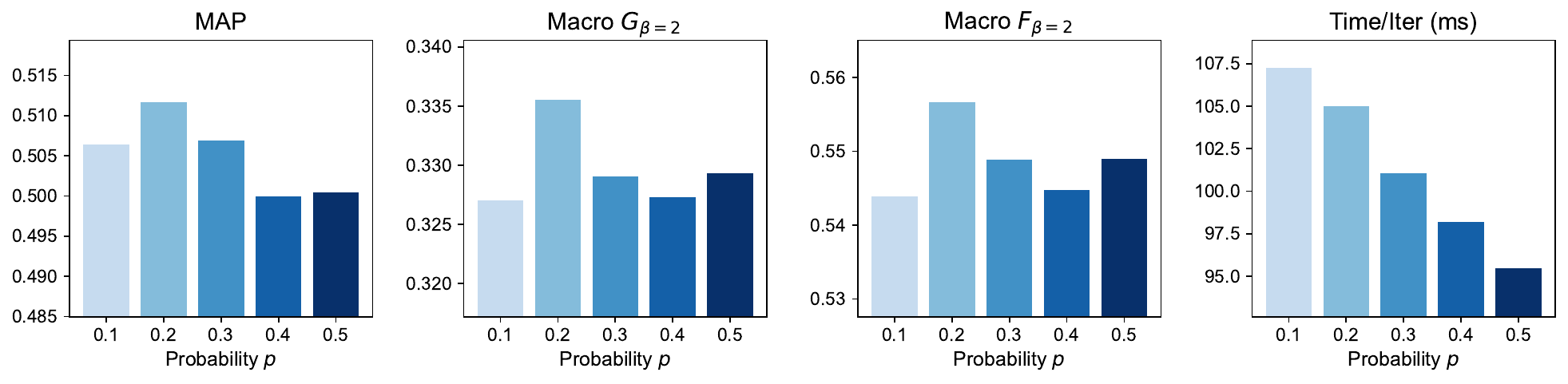}
\end{center}
\caption{Effect of the deactivation probability.Averaged performance and training time of the CE-SSL across four datasets and six random seeds under different deactivation probability $p$.}
\label{fig:sens_drop}
\end{figure*}

For each pre-trained weight $W_0^i$ in the CE-SSL, the proposed RD-LoRA deactivates its low-rank matrices ($A^i, B^i$) in the current iteration at a probability of $p$, which produces multiple sub-networks during model training. All the low-rank matrices are activated in the testing stage, generating an ensemble network that combines all the sub-networks. Consequently, the probability $p$ is an important parameter that controls the training time and the final performance of the proposed CE-SSL. In Fig.\ref{fig:sens_drop}, we adjust $p$ from 0.1 to 0.5 and present the averaged model performance across four datasets, including the training time for each iteration. Note that the labeled ratio is set to 5\%, and the initial ranks for all the low-rank matrices are set to 16. The results show that the CE-SSL with $p=0.2$ demonstrates the best detection performance compared with the model with other settings. In addition, it can be observed that the training time of the CE-SSL decreases as $p$ increases. The reason is that the larger the $p$ is, the more low-rank matrices are deactivated during model training, which speeds up the forward-backward propagation.

\subsection{Performance Comparisons under Various Backbone Sizes}
In the previous sections, we have already proved the robustness and computation efficiency of the proposed CE-SSL under a base backbone with 9.505 million parameters. Here, we compare its performance with other baseline models under medium and large backbones, which share the same architecture as the base backbone but have more parameters (Table \ref{tab:settings}). Specifically, the medium backbone has 50.494 million parameters, and the large backbone has 113.490 million parameters. They are pre-trained on the CODE-full dataset, a huge but restricted ECG dataset with 2,322,513 ECG recordings from 1,558,772 patients\cite{ribeiro2019tele,ribeiro2020automatic}. In Supplementary Materials Table S9 and Table S10, we report the performance of CE-SSL and semi-supervised baselines on the medium and the large backbones, respectively. The results demonstrate that CE-SSL achieves similar and even better CVDs detection performance than the semi-supervised baselines and exhibits the lowest computation costs. For example, using the medium backbone, CE-SSL achieves a macro $F_{\beta=2}$ of 0.599$\pm$0.010, which is 3.7\% larger than the second-best model's (SAW) performance in the PTB-XL dataset. Using the large backbone, CE-SSL achieves a macro $F_{\beta=2}$ of 0.565$\pm$0.010 in the G12EC dataset, outperforming SAW by 3.1\%. Regarding the computational costs, the number of trainable parameters of CE-SSL is 0.9\% to 3.1\% of the other baselines on the medium backbone and 0.6\% to 2.1\% on the large backbone. In addition, CE-SSL demonstrates the lowest GPU memory consumption and the highest training speed compared to the other semi-supervised baselines. For the memory footprint, CE-SSL achieves an average GPU memory usage of 6.16 GB using the medium backbone and 9.22 GB using the large backbone, 3.09 GB and 4.59 GB less than the second-best model (Adsh). Furthermore, CE-SSL achieves an average training time per iteration of 259.25 ms using the medium backbone and 485.5 ms using the large backbone, 162.5 ms and 289.75 ms faster than the second-best model (MixedTeacher). These phenomenons demonstrate that as the number of model parameters increases, the computational efficiency advantage of CE-SSL over other models becomes increasingly apparent.
In Supplementary Materials Table S11 and Table S12, we present the performance of CE-SSL and parameter-efficient semi-supervised methods on the medium and large backbones, respectively. It can be observed that CE-SSL outperforms the other models in CVDs detection on both medium and large backbones. Additionally, CE-SSL demonstrates the fastest training speed across four datasets compared with other parameter-efficient methods. In Supplementary Materials Section 3 Fig. S2-S3, we provide the paired t-test results of the model performance on the two backbones. The statistical results indicate that CE-SSL outperforms the above baselines in ECG-based CVDs detection at a 0.05 significance level in most conditions. 

\subsection{Toward Higher Computational Efficiency in Clinical Practices}
\begin{figure*}[p]
  \begin{center}
  \subfloat[The peak GPU memory footprints during training]{\includegraphics[width=1\textwidth]{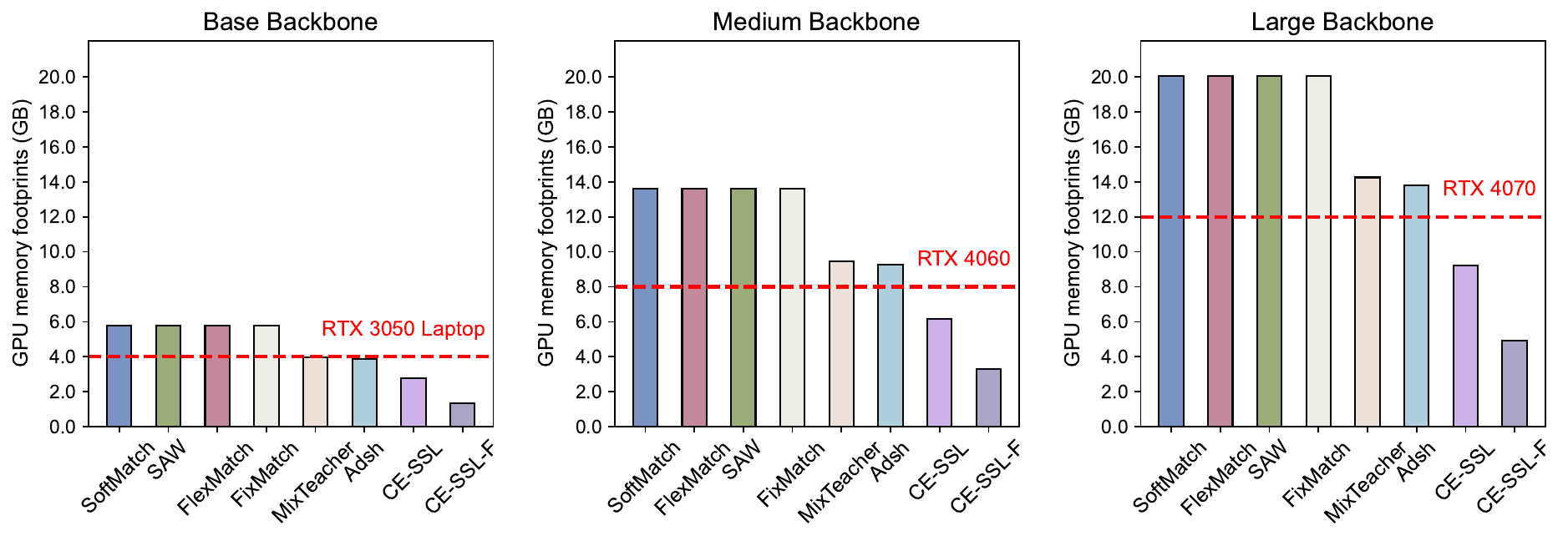}\label{fig:freeze_memory}}\\
  \subfloat[The average training time for each optimization iteration]{\includegraphics[width=1\textwidth]{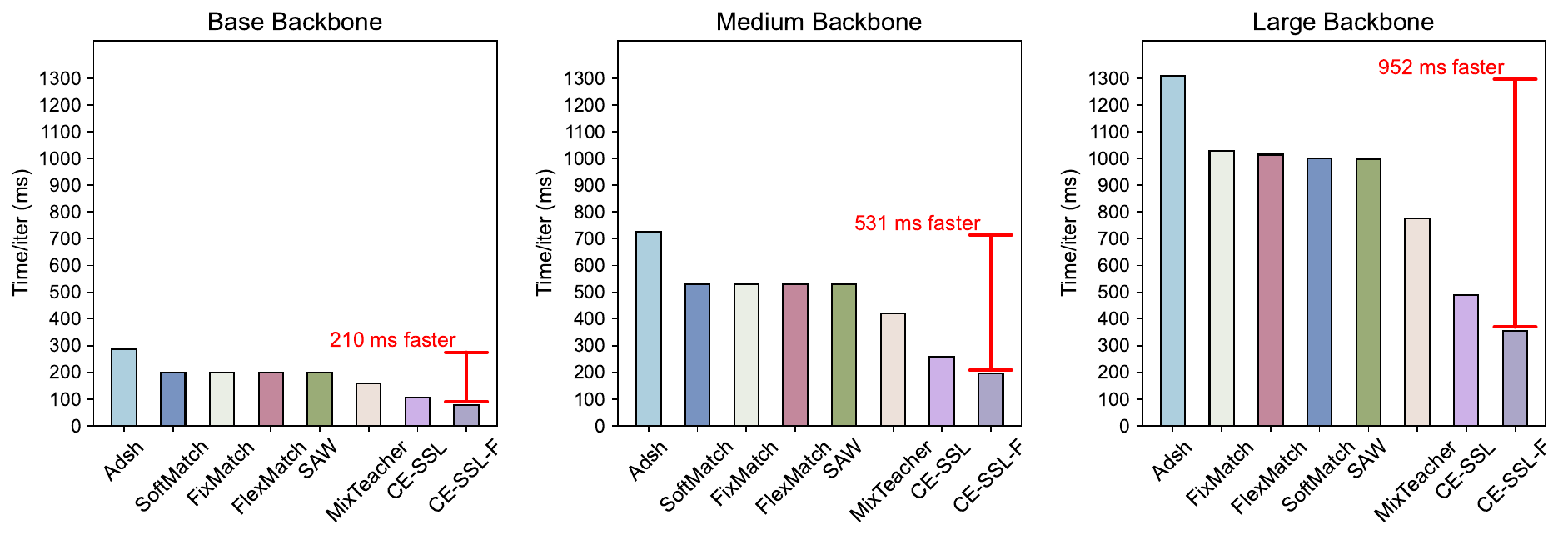}\label{fig:freeze_runningtime}}\\
  \subfloat[Macro $F_{\beta=2}$ score]{\includegraphics[width=1\textwidth]{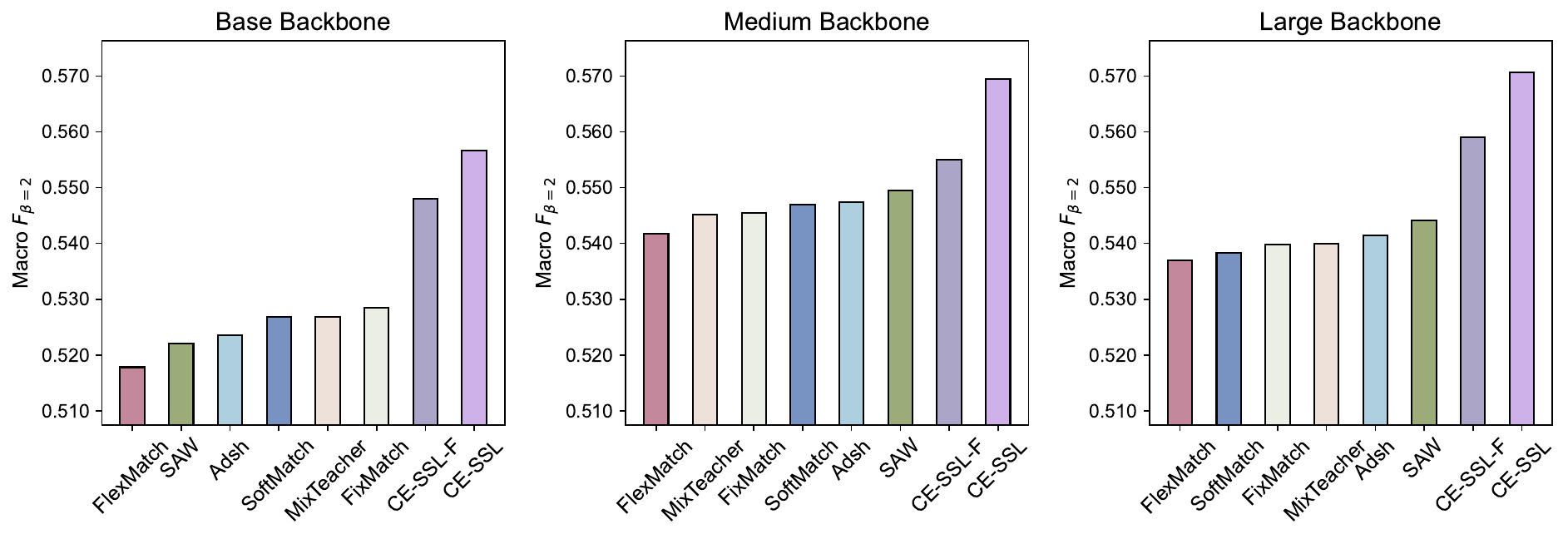}\label{fig:freeze_Fbeta}}
  \end{center}
  \caption{Evaluation of the average detection performance and computational efficiency of various semi-supervised methods for CVDs across four downstream datasets.  Specifically, the macro $F_{\beta=2}$ scores are used to evaluate their CVDs detection performance. Additionally, the average training time per optimization iteration and the maximum GPU memory usage are presented to evaluate computational efficiency. Methods exceeding the GPU memory thresholds, indicated by the red dashed lines, are not deployable on the corresponding NVIDIA GPU cards. The only distinction between CE-SSL-F and CE-SSL lies in the training process. Specifically, CE-SSL-F freezes the first two convolution blocks during training, whereas CE-SSL does not impose such constraints on them. The initial rank for CE-SSL and CE-SSL-F is set to 16 and 4, respectively.}
  \label{fig:freeze}
\end{figure*}

Although deploying the CE-SSL paradigm with the base backbone on low-level devices (4-6 GB GPU memory) is easy, implementing the paradigm with the medium and large backbones is still challenging. To overcome this limitation, we adopt a simple but effective approach to boost the computational efficiency of the CE-SSL. Specifically, we freeze the first two convolution blocks in the backbones during the CE-SSL training process. The new paradigm is denoted as 'CE-SSL-F' in the following analysis. We present the CVDs detection performance and the computational efficiency of CE-SSL-F, CE-SSL, and the SOTA methods in semi-supervised learning in Fig.\ref{fig:freeze}. Note that the batch sizes for all the compared methods are set to 64. 

First, freezing the convolution blocks greatly reduces the cached activation during the forward pass, significantly decreasing the GPU memory footprints. As shown in Fig.\ref{fig:freeze_memory}, it can be observed that the CE-SSL-F requires nearly 50\% less GPU memory footprints compared to the CE-SSL, generalizing its applications in low-level devices (NVIDIA RTX 3050 laptops and RTX 4060 GPU cards). Specifically, CE-SSL-F is deployable on RTX 3050 laptops with both base and medium backbones, and it is the only method that can be implemented on the RTX 4060 GPU cards with a large backbone. In contrast, deploying the CE-SSL with a large backbone requires medium-level devices (NVIDIA RTX 4070 GPU cards), while other semi-supervised methods require high-level devices with GPU memory larger than 12 GB. Second, the parameters of the frozen blocks are not updated during the backward pass, which increases the training speed of CE-SSL-F. The larger the backbone is, the more parameters are frozen, and thus the more gradient backward time is saved. As shown in Fig.\ref{fig:freeze_runningtime}, CE-SSL-F demonstrates the fastest training speed compared with other models, and its advantages become more significant along with the increase in backbone sizes. Third, CE-SSL-F only sacrifices 1-2\% CVDs detection performance compared with CE-SSL. More importantly, it consistently outperforms the other semi-supervised methods across different backbones (Fig.\ref{fig:freeze_Fbeta}), demonstrating its effectiveness in CVDs detection. This phenomenon can be explained by the strong transferability of the pre-trained convolution blocks located in the first few layers of the backbone\cite{sharif2014cnn,tajbakhsh2016convolutional}. Specifically, they mainly contain domain-invariant knowledge for CVDs detection, and their parameters will not be changed significantly during the fine-tuning process. Therefore, freezing them does not greatly decrease the model performance.

In summary, the experiment results illustrate that the computational efficiency of the CE-SSL can be increased to adapt to low-level devices without losing its superior CVDs detection performance compared to other semi-supervised methods. This advantage demonstrates CE-SSL's flexibility in different clinical application scenarios with various computational resources. 
\subsection{Ablation Study}
We conducted an ablation study to evaluate the contribution of the modules implemented in CE-SSL. Specifically, we remove each of the proposed three modules from the CE-SSL and record the corresponding model performance on the four datasets. Note that the initial rank $r$ is 16 for all the compared models. \textbf{(1) The random-deactivation low-rank adaptation improves model performance and computational efficiency.} It randomly deactivates the low-rank matrices in each trainable layer with a given probability, which generates multiple sub-networks during model training. All the low-rank matrices are activated in the testing stage, and the sub-networks are merged into a robust ensemble network. In the G12EC dataset, the CE-SSL with the random deactivation module increases the macro $F_{\beta=2}$ from 0.536$\pm$0.021 to 0.551$\pm$0.017 and decreases the ranking loss from 0.095$\pm$0.004 to 0.092$\pm$0.002. Additionally, the Time/iter is reduced from 104 ms to 98 ms. This observation indicates that the computation cost of training a sub-network with fewer low-rank matrices is lower than optimizing the entire network. \textbf{(2) The one-shot rank allocation method improves low-rank adaptation performance with high computational efficiency.} One drawback of LoRA\cite{hu2022lora} is its inability to allocate an optimal rank for each incremental matrix\cite{zhang2023adaptive,zhang2023increlora}. Here, the proposed rank allocation method determines the optimal ranks using only one gradient-backward iteration with high computational efficiency. As shown in Table \ref{tab:Ablation}, removing the proposed module decreases the detection performance of CE-SSL in the four datasets. For example, the macro $F_{\beta=2}$ decreases from 0.530$\pm$0.008 to 0.514$\pm$0.018, and the coverage increases from 2.483$\pm$0.055 to 2.503$\pm$0.040 on the Chapman dataset. Additionally, it can be observed that the one-shot rank allocation module does not introduce heavy computational burdens (Time/iter only increases by 1-2ms), demonstrating its high computational efficiency. (3) \textbf{The proposed lightweight semi-supervised learning benefits model performance under limited supervision without greatly increasing the training time.} It utilizes the unlabeled data to stabilize the statistics within the BN layers in the convolution blocks, preventing them from over-fitting to small amounts of labeled data. When the module is removed, the performance of the CE-SSL decreases on all the datasets. For example, the macro $F_{\beta=2}$ score decreases from 0.551$\pm$0.017 to  0.536$\pm$0.029 on the G12EC dataset. Compared to other semi-supervised baselines (Table \ref{tab:compare_semi}), the extra computation costs (Time/iter only increases by 20ms) caused by the proposed module are much lower. In Supplementary Materials Section 3, ablation studies on the medium and large backbones are provided, consistently demonstrating the notable contribution of the proposed three techniques.
\begin{table*}[t]
\fontsize{8}{11}\selectfont
\setlength{\tabcolsep}{0.1em}
\begin{center}
\caption{Ablation study of the proposed CE-SSL on the base backbone. 'w/o random deactivation' represents the CE-SSL without the random deactivation technique, and the deactivation probability $p$ is set to zero. 'w/o rank allocation' represents the CE-SSL without the one-shot rank allocation, and all pre-trained weights are updated with the initial rank $r=16$. 'w/o semi-supervised BN' denotes the CE-SSL without the semi-supervised batch normalization for lightweight semi-supervised learning.}
\label{tab:Ablation}
\scalebox{1}{
\color{black}
\begin{tabular*}{\hsize}{@{}@{\extracolsep{\fill}}lccccccc@{}}
\toprule
Methods & Time/iter $\downarrow$ & Ranking Loss $\downarrow$ & Coverage $\downarrow$ & Macro AUC $\uparrow$ & MAP $\uparrow$ & Macro $G_{\beta=2}$ $\uparrow$  & Macro $F_{beta}$ $\uparrow$\\
\midrule
\multicolumn{8}{c}{\textbf{G12EC Dataset}}\\
\midrule
w/o random deactivation & 104ms&0.095$\pm$0.004&3.954$\pm$0.163&0.848$\pm$0.007&0.470$\pm$0.007&0.294$\pm$0.015&0.536$\pm$0.021\\
w/o rank allocation &97ms&0.092$\pm$0.002&3.848$\pm$0.049&0.849$\pm$0.007&0.467$\pm$0.009&0.294$\pm$0.016&0.537$\pm$0.019\\
w/o semi-supervised BN & 78ms&0.092$\pm$0.002&3.895$\pm$0.104&0.854$\pm$0.004&0.475$\pm$0.011&0.297$\pm$0.021&0.536$\pm$0.029\\
\textbf{CE-SSL} & \textbf{98ms}&\textbf{0.092$\pm$0.002}&\textbf{3.867$\pm$0.088}&\textbf{0.855$\pm$0.005}&\textbf{0.476$\pm$0.006}&\textbf{0.307$\pm$0.016}&\textbf{0.551$\pm$0.017}\\
\midrule
\multicolumn{8}{c}{\textbf{PTB-XL Dataset}}\\
\midrule
w/o random deactivation & 115ms&0.034$\pm$0.002&2.741$\pm$0.062&0.890$\pm$0.005&0.516$\pm$0.009&0.328$\pm$0.012&0.554$\pm$0.011\\
w/o rank allocation &108ms&0.032$\pm$0.001&2.692$\pm$0.046&0.895$\pm$0.003&0.530$\pm$0.005&0.332$\pm$0.011&0.560$\pm$0.014\\
w/o semi-supervised BN & 87ms&0.031$\pm$0.002&2.670$\pm$0.064&0.899$\pm$0.004&0.532$\pm$0.006&0.332$\pm$0.010&0.565$\pm$0.007\\
\textbf{CE-SSL} & \textbf{110ms}&\textbf{0.031$\pm$0.000}&\textbf{2.641$\pm$0.020}&\textbf{0.901$\pm$0.003}&\textbf{0.530$\pm$0.005}&\textbf{0.346$\pm$0.006}&\textbf{0.578$\pm$0.006}\\
\midrule
\multicolumn{8}{c}{\textbf{Ningbo Dataset}}\\
\midrule
w/o random deactivation & 121ms&0.032$\pm$0.003&2.887$\pm$0.085&0.925$\pm$0.005&0.497$\pm$0.015&0.321$\pm$0.013&0.553$\pm$0.017\\
w/o rank allocation &114ms&0.030$\pm$0.001&2.801$\pm$0.023&0.928$\pm$0.002&0.497$\pm$0.021&0.325$\pm$0.010&0.563$\pm$0.014\\
w/o semi-supervised BN & 92ms&0.031$\pm$0.001&2.821$\pm$0.058&0.929$\pm$0.003&0.499$\pm$0.017&0.325$\pm$0.012&0.559$\pm$0.018\\
\textbf{CE-SSL} & \textbf{115ms}&\textbf{0.030$\pm$0.001}&\textbf{2.805$\pm$0.063}&\textbf{0.928$\pm$0.002}&\textbf{0.505$\pm$0.019}&\textbf{0.334$\pm$0.011}&\textbf{0.569$\pm$0.014}\\
\midrule
\multicolumn{8}{c}{\textbf{Chapman Dataset}}\\
\midrule
w/o random deactivation & 102ms&0.041$\pm$0.003&2.505$\pm$0.080&0.895$\pm$0.010&0.526$\pm$0.005&0.335$\pm$0.012&0.514$\pm$0.015\\
w/o rank allocation &96ms&0.041$\pm$0.001&2.503$\pm$0.040&0.892$\pm$0.008&0.527$\pm$0.012&0.346$\pm$0.007&0.514$\pm$0.018\\
w/o semi-supervised BN & 77ms&0.040$\pm$0.002&2.468$\pm$0.050&0.896$\pm$0.010&0.533$\pm$0.010&0.350$\pm$0.020&0.527$\pm$0.026\\
\textbf{CE-SSL} & \textbf{97ms}&\textbf{0.040$\pm$0.002}&\textbf{2.483$\pm$0.055}&\textbf{0.896$\pm$0.006}&\textbf{0.536$\pm$0.004}&\textbf{0.355$\pm$0.005}&\textbf{0.530$\pm$0.008}\\
\bottomrule
\end{tabular*}
}
\end{center}
\end{table*}

\section{Conclusion}
Bottlenecks in model performance and computational efficiency have become great challenges in the clinical application of CVDs detection systems based on pre-trained models, especially when the supervised information is scarce in the downstream ECG datasets. Previous studies usually overcome the performance bottleneck at the cost of a large drop in computational efficiency\cite{selftune2021,zhang2023adaptive,zhang2023increlora,peiris2023uncertainty}. In this paper, we propose a computationally efficient semi-supervised learning paradigm (CE-SSL) for adapting pre-trained models on downstream datasets with limited supervision and high computational efficiency. Experiment results on four downstream ECG datasets and three backbone settings indicate that CE-SSL achieves superior CVDs detection performance and computational efficiency compared to state-of-the-art methods. In conclusion, our study offers a fast and robust semi-supervised learning paradigm for ECG-based CVDs detection under limited supervision. It provides a feasible solution for efficiently adapting pre-trained models on downstream ECG datasets. We hope this learning paradigm will pave the way for the application of automatic CVDs detection systems and broaden its applicability to various ECG-based tasks. 

\section{Limitation and Future Work}
Apart from the label scarcity problem, the class imbalance problem also limits the model performance on ECG recordings with rare CVDs. Specifically, it tends to make negative predictions on the rare CVDs to achieve minimal training loss, which might result in sub-optimal classification performance on the test data with different class distributions. As such, developing a robust method for multi-label CVDs classification under imbalance class distributions is still an ongoing issue that deserves more attention from future studies.

%

\ifCLASSOPTIONcompsoc
  \section*{Acknowledgments}
\else
  \section*{Acknowledgment}
\fi

This work was supported by an RAEng Research Chair, the InnoHK Hong Kong projects under the Hong Kong Center for Cerebro-cardiovascular Health Engineering (COCHE), the National Natural Science Foundation of China (22322816) and the City University of Hong Kong Project (9610640).
\bibliographystyle{IEEEtran}
\bibliography{references}

\begin{thebibliography}{10}
\providecommand{\url}[1]{#1}
\csname url@samestyle\endcsname
\providecommand{\newblock}{\relax}
\providecommand{\bibinfo}[2]{#2}
\providecommand{\BIBentrySTDinterwordspacing}{\spaceskip=0pt\relax}
\providecommand{\BIBentryALTinterwordstretchfactor}{4}
\providecommand{\BIBentryALTinterwordspacing}{\spaceskip=\fontdimen2\font plus
\BIBentryALTinterwordstretchfactor\fontdimen3\font minus \fontdimen4\font\relax}
\providecommand{\BIBforeignlanguage}[2]{{%
\expandafter\ifx\csname l@#1\endcsname\relax
\typeout{** WARNING: IEEEtran.bst: No hyphenation pattern has been}%
\typeout{** loaded for the language `#1'. Using the pattern for}%
\typeout{** the default language instead.}%
\else
\language=\csname l@#1\endcsname
\fi
#2}}
\providecommand{\BIBdecl}{\relax}
\BIBdecl

\bibitem{kelly2010promoting}
B.~B. Kelly, V.~Fuster \emph{et~al.}, \emph{Promoting cardiovascular health in the developing world: a critical challenge to achieve global health}.\hskip 1em plus 0.5em minus 0.4em\relax National Academies Press, 2010.

\bibitem{kiyasseh2021clinical}
D.~Kiyasseh, T.~Zhu, and D.~Clifton, ``A clinical deep learning framework for continually learning from cardiac signals across diseases, time, modalities, and institutions,'' \emph{Nature Communications}, vol.~12, no.~1, p. 4221, 2021.

\bibitem{lai2023practical}
J.~Lai, H.~Tan, J.~Wang, L.~Ji, J.~Guo, B.~Han, Y.~Shi, Q.~Feng, and W.~Yang, ``Practical intelligent diagnostic algorithm for wearable 12-lead {ECG} via self-supervised learning on large-scale dataset,'' \emph{Nature Communications}, vol.~14, no.~1, p. 3741, 2023.

\bibitem{hannun2019cardiologist}
A.~Y. Hannun, P.~Rajpurkar, M.~Haghpanahi, G.~H. Tison, C.~Bourn, M.~P. Turakhia, and A.~Y. Ng, ``Cardiologist-level arrhythmia detection and classification in ambulatory electrocardiograms using a deep neural network,'' \emph{Nature medicine}, vol.~25, no.~1, pp. 65--69, 2019.

\bibitem{ribeiro2020automatic}
A.~H. Ribeiro, M.~H. Ribeiro, G.~M. Paix{\~a}o, D.~M. Oliveira, P.~R. Gomes, J.~A. Canazart, M.~P. Ferreira, C.~R. Andersson, P.~W. Macfarlane, W.~Meira~Jr \emph{et~al.}, ``Automatic diagnosis of the 12-lead {ECG} using a deep neural network,'' \emph{Nature communications}, vol.~11, no.~1, p. 1760, 2020.

\bibitem{al2023machine}
S.~S. Al-Zaiti, C.~Martin-Gill, J.~K. Z{\`e}gre-Hemsey, Z.~Bouzid, Z.~Faramand, M.~O. Alrawashdeh, R.~E. Gregg, S.~Helman, N.~T. Riek, K.~Kraevsky-Phillips \emph{et~al.}, ``Machine learning for ecg diagnosis and risk stratification of occlusion myocardial infarction,'' \emph{Nature Medicine}, vol.~29, no.~7, pp. 1804--1813, 2023.

\bibitem{lu2024decoding}
L.~Lu, T.~Zhu, A.~H. Ribeiro, L.~Clifton, E.~Zhao, J.~Zhou, A.~L.~P. Ribeiro, Y.-T. Zhang, and D.~A. Clifton, ``Decoding 2.3 million ecgs: interpretable deep learning for advancing cardiovascular diagnosis and mortality risk stratification,'' \emph{European Heart Journal-Digital Health}, vol.~5, no.~3, pp. 247--259, 2024.

\bibitem{berthelot2019mixmatch}
D.~Berthelot, N.~Carlini, I.~Goodfellow, N.~Papernot, A.~Oliver, and C.~A. Raffel, ``Mixmatch: A holistic approach to semi-supervised learning,'' 2019.

\bibitem{sohn2020fixmatch}
K.~Sohn, D.~Berthelot, N.~Carlini, Z.~Zhang, H.~Zhang, C.~A. Raffel, E.~D. Cubuk, A.~Kurakin, and C.-L. Li, ``Fixmatch: Simplifying semi-supervised learning with consistency and confidence,'' pp. 596--608, 2020.

\bibitem{zhang2022semi}
P.~Zhang, Y.~Chen, F.~Lin, S.~Wu, X.~Yang, and Q.~Li, ``Semi-supervised learning for automatic atrial fibrillation detection in 24-hour holter monitoring,'' \emph{IEEE Journal of Biomedical and Health Informatics}, vol.~26, no.~8, pp. 3791--3801, 2022.

\bibitem{zhoupami2023}
R.~Zhou, L.~Lu, Z.~Liu, T.~Xiang, Z.~Liang, D.~A. Clifton, Y.~Dong, and Y.-T. Zhang, ``Semi-supervised learning for multi-label cardiovascular diseases prediction: A multi-dataset study,'' \emph{IEEE Transactions on Pattern Analysis and Machine Intelligence}, pp. 1--17, 2023.

\bibitem{vaswani2017attention}
A.~Vaswani, N.~Shazeer, N.~Parmar, J.~Uszkoreit, L.~Jones, A.~N. Gomez, {\L}.~Kaiser, and I.~Polosukhin, ``Attention is all you need,'' \emph{Advances in neural information processing systems}, vol.~30, 2017.

\bibitem{radford2019language}
A.~Radford, J.~Wu, R.~Child, D.~Luan, D.~Amodei, and I.~Sutskever, ``Language models are unsupervised multitask learners,'' 2019.

\bibitem{he2022masked}
K.~He, X.~Chen, S.~Xie, Y.~Li, P.~Doll{\'a}r, and R.~Girshick, ``Masked autoencoders are scalable vision learners,'' pp. 16\,000--16\,009, 2022.

\bibitem{selftune2021}
\BIBentryALTinterwordspacing
X.~Wang, J.~Gao, M.~Long, and J.~Wang, ``Self-tuning for data-efficient deep learning,'' in \emph{Proceedings of the 38th International Conference on Machine Learning}, ser. Proceedings of Machine Learning Research, M.~Meila and T.~Zhang, Eds., vol. 139.\hskip 1em plus 0.5em minus 0.4em\relax PMLR, 18--24 Jul 2021, pp. 10\,738--10\,748. [Online]. Available: \url{https://proceedings.mlr.press/v139/wang21g.html}
\BIBentrySTDinterwordspacing

\bibitem{zhou2018semi}
H.-Y. Zhou, A.~Oliver, J.~Wu, and Y.~Zheng, ``When semi-supervised learning meets transfer learning: Training strategies, models and datasets,'' \emph{arXiv preprint arXiv:1812.05313}, 2018.

\bibitem{comatch2021}
J.~Li, C.~Xiong, and S.~C. Hoi, ``Comatch: Semi-supervised learning with contrastive graph regularization,'' pp. 9475--9484, October 2021.

\bibitem{zhang2021flexmatch}
B.~Zhang, Y.~Wang, W.~Hou, H.~WU, J.~Wang, M.~Okumura, and T.~Shinozaki, ``Flexmatch: Boosting semi-supervised learning with curriculum pseudo labeling,'' pp. 18\,408--18\,419, 2021.

\bibitem{peiris2023uncertainty}
H.~Peiris, M.~Hayat, Z.~Chen, G.~Egan, and M.~Harandi, ``Uncertainty-guided dual-views for semi-supervised volumetric medical image segmentation,'' \emph{Nature Machine Intelligence}, vol.~5, no.~7, pp. 724--738, 2023.

\bibitem{berthelot2019remixmatch}
D.~Berthelot, N.~Carlini, E.~D. Cubuk, A.~Kurakin, K.~Sohn, H.~Zhang, and C.~Raffel, ``Remixmatch: Semi-supervised learning with distribution matching and augmentation anchoring,'' 2020.

\bibitem{chen2023softmatch}
H.~Chen, R.~Tao, Y.~Fan, Y.~Wang, J.~Wang, B.~Schiele, X.~Xie, B.~Raj, and M.~Savvides, ``Softmatch: Addressing the quantity-quality tradeoff in semi-supervised learning,'' 2023.

\bibitem{vaid2023foundational}
A.~Vaid, J.~Jiang, A.~Sawant, S.~Lerakis, E.~Argulian, Y.~Ahuja, J.~Lampert, A.~Charney, H.~Greenspan, J.~Narula \emph{et~al.}, ``A foundational vision transformer improves diagnostic performance for electrocardiograms,'' \emph{NPJ Digital Medicine}, vol.~6, no.~1, p. 108, 2023.

\bibitem{han2024foundation}
Y.~Han and C.~Ding, ``Foundation models in electrocardiogram: A review,'' \emph{arXiv preprint arXiv:2410.19877}, 2024.

\bibitem{mathew2024foundation}
G.~Mathew, D.~Barbosa, J.~Prince, and S.~Venkatraman, ``Foundation models for cardiovascular disease detection via biosignals from digital stethoscopes,'' \emph{npj Cardiovascular Health}, vol.~1, no.~1, p.~25, 2024.

\bibitem{mckeen2024ecg}
K.~McKeen, L.~Oliva, S.~Masood, A.~Toma, B.~Rubin, and B.~Wang, ``Ecg-fm: An open electrocardiogram foundation model,'' \emph{arXiv preprint arXiv:2408.05178}, 2024.

\bibitem{pham2024c}
M.~Pham, A.~Saeed, and D.~Ma, ``C-melt: Contrastive enhanced masked auto-encoders for ecg-language pre-training,'' \emph{arXiv preprint arXiv:2410.02131}, 2024.

\bibitem{hu2022lora}
\BIBentryALTinterwordspacing
E.~J. Hu, Y.~Shen, P.~Wallis, Z.~Allen-Zhu, Y.~Li, S.~Wang, L.~Wang, and W.~Chen, ``Lo{RA}: Low-rank adaptation of large language models,'' 2022. [Online]. Available: \url{https://openreview.net/forum?id=nZeVKeeFYf9}
\BIBentrySTDinterwordspacing

\bibitem{zhang2023adaptive}
\BIBentryALTinterwordspacing
Q.~Zhang, M.~Chen, A.~Bukharin, P.~He, Y.~Cheng, W.~Chen, and T.~Zhao, ``Adaptive budget allocation for parameter-efficient fine-tuning,'' 2023. [Online]. Available: \url{https://openreview.net/forum?id=lq62uWRJjiY}
\BIBentrySTDinterwordspacing

\bibitem{houlsby2019parameter}
N.~Houlsby, A.~Giurgiu, S.~Jastrzebski, B.~Morrone, Q.~De~Laroussilhe, A.~Gesmundo, M.~Attariyan, and S.~Gelly, ``Parameter-efficient transfer learning for nlp,'' PMLR, pp. 2790--2799, 2019.

\bibitem{zaken2021bitfit}
E.~B. Zaken, S.~Ravfogel, and Y.~Goldberg, ``Bitfit: Simple parameter-efficient fine-tuning for transformer-based masked language-models,'' \emph{arXiv preprint arXiv:2106.10199}, 2021.

\bibitem{chen2023hadamard}
Y.~Chen, Q.~Fu, G.~Fan, L.~Du, J.-G. Lou, S.~Han, D.~Zhang, Z.~Li, and Y.~Xiao, ``Hadamard adapter: An extreme parameter-efficient adapter tuning method for pre-trained language models,'' pp. 276--285, 2023.

\bibitem{zhang2023increlora}
F.~Zhang, L.~Li, J.~Chen, Z.~Jiang, B.~Wang, and Y.~Qian, ``Increlora: Incremental parameter allocation method for parameter-efficient fine-tuning,'' \emph{arXiv preprint arXiv:2308.12043}, 2023.

\bibitem{ding2023parameter}
N.~Ding, Y.~Qin, G.~Yang, F.~Wei, Z.~Yang, Y.~Su, S.~Hu, Y.~Chen, C.-M. Chan, W.~Chen \emph{et~al.}, ``Parameter-efficient fine-tuning of large-scale pre-trained language models,'' \emph{Nature Machine Intelligence}, vol.~5, no.~3, pp. 220--235, 2023.

\bibitem{Pour2018}
B.~Pourbabaee, M.~J. Roshtkhari, and K.~Khorasani, ``Deep convolutional neural networks and learning ecg features for screening paroxysmal atrial fibrillation patients,'' \emph{IEEE Transactions on Systems, Man, and Cybernetics: Systems}, vol.~48, no.~12, pp. 2095--2104, 2018.

\bibitem{strodthoff2020deep}
N.~Strodthoff, P.~Wagner, T.~Schaeffter, and W.~Samek, ``Deep learning for {ECG} analysis: Benchmarks and insights from {PTB-XL},'' \emph{IEEE Journal of Biomedical and Health Informatics}, vol.~25, no.~5, pp. 1519--1528, 2020.

\bibitem{kiyasseh2021clocs}
D.~Kiyasseh, T.~Zhu, and D.~A. Clifton, ``Clocs: Contrastive learning of cardiac signals across space, time, and patients,'' PMLR, pp. 5606--5615, 2021.

\bibitem{huang2022snippet}
Y.~Huang, G.~G. Yen, and V.~S. Tseng, ``Snippet policy network v2: Knee-guided neuroevolution for multi-lead {ECG} early classification,'' \emph{IEEE Transactions on Neural Networks and Learning Systems}, 2022.

\bibitem{2023dylora}
\BIBentryALTinterwordspacing
M.~Valipour, M.~Rezagholizadeh, I.~Kobyzev, and A.~Ghodsi, ``{D}y{L}o{RA}: Parameter-efficient tuning of pre-trained models using dynamic search-free low-rank adaptation,'' in \emph{Proceedings of the 17th Conference of the European Chapter of the Association for Computational Linguistics}, A.~Vlachos and I.~Augenstein, Eds.\hskip 1em plus 0.5em minus 0.4em\relax Dubrovnik, Croatia: Association for Computational Linguistics, May 2023, pp. 3274--3287. [Online]. Available: \url{https://aclanthology.org/2023.eacl-main.239}
\BIBentrySTDinterwordspacing

\bibitem{ribeiro2019tele}
A.~L.~P. Ribeiro, G.~M. Paixao, P.~R. Gomes, M.~H. Ribeiro, A.~H. Ribeiro, J.~A. Canazart, D.~M. Oliveira, M.~P. Ferreira, E.~M. Lima, J.~L. de~Moraes \emph{et~al.}, ``Tele-electrocardiography and bigdata: the code (clinical outcomes in digital electrocardiography) study,'' \emph{Journal of electrocardiology}, vol.~57, pp. S75--S78, 2019.

\bibitem{Leilu2024}
\BIBentryALTinterwordspacing
L.~Lu, T.~Zhu, A.~H. Ribeiro, L.~Clifton, E.~Zhao, J.~Zhou, A.~L.~P. Ribeiro, Y.-T. Zhang, and D.~A. Clifton, ``{Decoding 2.3 million ECGs: interpretable deep learning for advancing cardiovascular diagnosis and mortality risk stratification},'' \emph{European Heart Journal - Digital Health}, p. ztae014, 02 2024. [Online]. Available: \url{https://doi.org/10.1093/ehjdh/ztae014}
\BIBentrySTDinterwordspacing

\bibitem{AIMTbackbone}
P.~Nejedly, A.~Ivora, R.~Smisek, I.~Viscor, Z.~Koscova, P.~Jurak, and F.~Plesinger, ``Classification of {ECG} using ensemble of residual {CNNs} with attention mechanism,'' IEEE, pp. 1--4, 2021.

\bibitem{zi2023delta}
B.~Zi, X.~Qi, L.~Wang, J.~Wang, K.-F. Wong, and L.~Zhang, ``Delta-lora: Fine-tuning high-rank parameters with the delta of low-rank matrices,'' \emph{arXiv preprint arXiv:2309.02411}, 2023.

\bibitem{molchanov2019taylor}
P.~Molchanov, A.~Mallya, S.~Tyree, I.~Frosio, and J.~Kautz, ``Importance estimation for neural network pruning,'' 2019.

\bibitem{semi2006}
O.~Chapelle, B.~Schölkopf, and A.~Zien, \emph{{Semi-Supervised Learning}}.\hskip 1em plus 0.5em minus 0.4em\relax The MIT Press, 09 2006, vol.~2.

\bibitem{yun2019cutmix}
S.~Yun, D.~Han, S.~J. Oh, S.~Chun, J.~Choe, and Y.~Yoo, ``Cutmix: Regularization strategy to train strong classifiers with localizable features,'' pp. 6023--6032, 2019.

\bibitem{alday2020classification}
E.~A.~P. Alday, A.~Gu, A.~J. Shah, C.~Robichaux, A.-K.~I. Wong, C.~Liu, F.~Liu, A.~B. Rad, A.~Elola, S.~Seyedi \emph{et~al.}, ``Classification of 12-lead {ECGs}: the physionet/computing in cardiology challenge 2020,'' \emph{Physiological measurement}, vol.~41, no.~12, p. 124003, 2020.

\bibitem{zheng202012}
J.~Zheng, J.~Zhang, S.~Danioko, H.~Yao, H.~Guo, and C.~Rakovski, ``A 12-lead electrocardiogram database for arrhythmia research covering more than 10,000 patients,'' \emph{Scientific data}, vol.~7, no.~1, p.~48, 2020.

\bibitem{zheng2020optimal}
J.~Zheng, H.~Chu, D.~Struppa, J.~Zhang, S.~M. Yacoub, H.~El-Askary, A.~Chang, L.~Ehwerhemuepha, I.~Abudayyeh, A.~Barrett \emph{et~al.}, ``Optimal multi-stage arrhythmia classification approach,'' \emph{Scientific reports}, vol.~10, no.~1, p. 2898, 2020.

\bibitem{wagner2020ptb}
P.~Wagner, N.~Strodthoff, R.-D. Bousseljot, D.~Kreiseler, F.~I. Lunze, W.~Samek, and T.~Schaeffter, ``{PTB-XL}, a large publicly available electrocardiography dataset,'' \emph{Scientific data}, vol.~7, no.~1, p. 154, 2020.

\bibitem{loshchilov2017decoupled}
I.~Loshchilov and F.~Hutter, ``Decoupled weight decay regularization,'' \emph{arXiv preprint arXiv:1711.05101}, 2017.

\bibitem{guo2022class}
L.-Z. Guo and Y.-F. Li, ``Class-imbalanced semi-supervised learning with adaptive thresholding,'' PMLR, pp. 8082--8094, 2022.

\bibitem{lai2022smoothed}
Z.~Lai, C.~Wang, H.~Gunawan, S.-C.~S. Cheung, and C.-N. Chuah, ``Smoothed adaptive weighting for imbalanced semi-supervised learning: Improve reliability against unknown distribution data,'' PMLR, pp. 11\,828--11\,843, 2022.

\bibitem{li2021prefix}
X.~L. Li and P.~Liang, ``Prefix-tuning: Optimizing continuous prompts for generation,'' \emph{arXiv preprint arXiv:2101.00190}, 2021.

\bibitem{sharif2014cnn}
A.~Sharif~Razavian, H.~Azizpour, J.~Sullivan, and S.~Carlsson, ``Cnn features off-the-shelf: an astounding baseline for recognition,'' in \emph{Proceedings of the IEEE conference on computer vision and pattern recognition workshops}, 2014, pp. 806--813.

\bibitem{tajbakhsh2016convolutional}
N.~Tajbakhsh, J.~Y. Shin, S.~R. Gurudu, R.~T. Hurst, C.~B. Kendall, M.~B. Gotway, and J.~Liang, ``Convolutional neural networks for medical image analysis: Full training or fine tuning?'' \emph{IEEE transactions on medical imaging}, vol.~35, no.~5, pp. 1299--1312, 2016.

\end{thebibliography}

\ifCLASSOPTIONcaptionsoff
  \newpage
\fi

\begin{IEEEbiography}[{\includegraphics[width=1in,height=1.25in,clip,keepaspectratio]{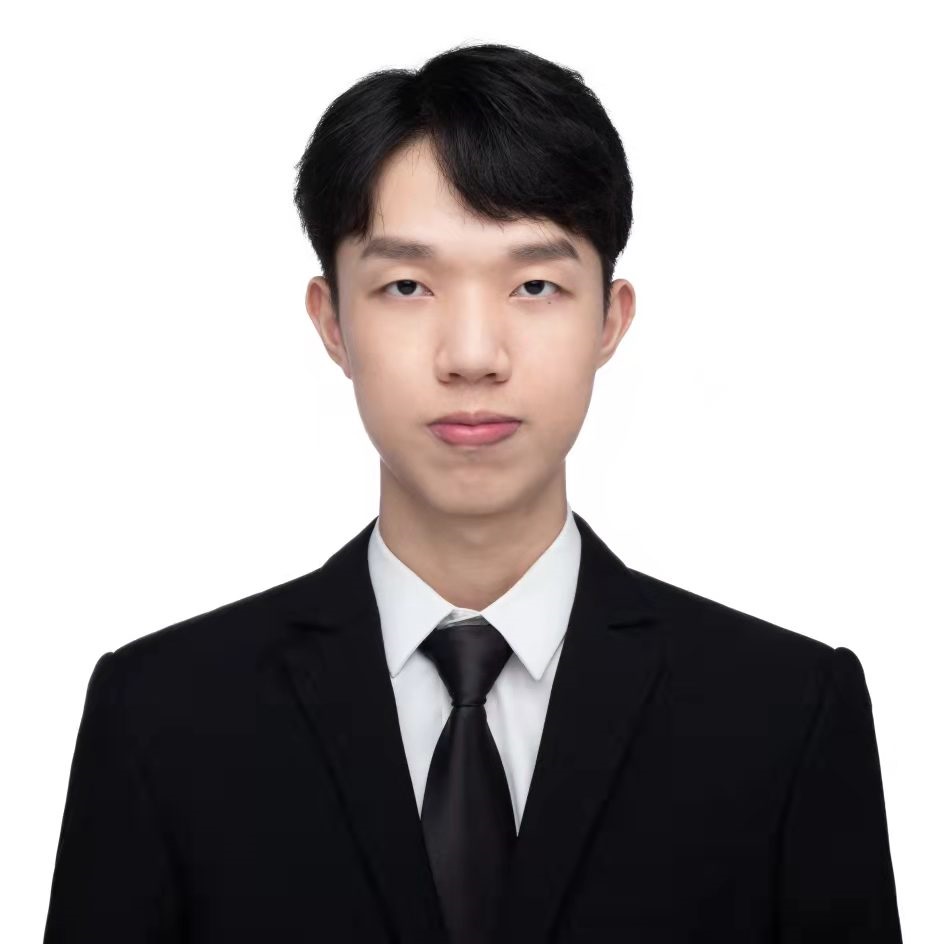}}]{Rushuang Zhou}
is a Ph.D. student in the Department of Biomedical Engineering, City University of Hong Kong, Hong Kong.
\end{IEEEbiography}

\begin{IEEEbiography}[{\includegraphics[width=1in,height=1.25in,clip,keepaspectratio]{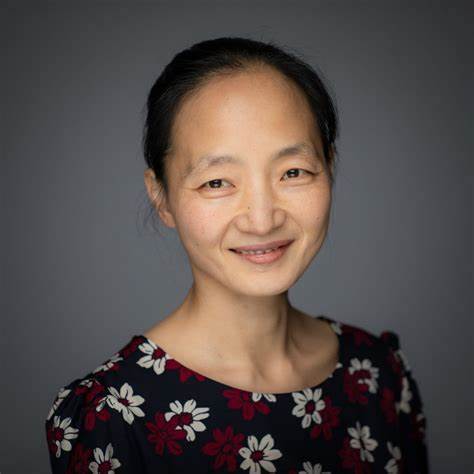}}]{Lei Clifton}
received the BSc and MSc degrees in electrical engineering from the Beijing Institute of Technology, and the PhD degree in statistical machine learning from the University of Manchester. She is currently the team leader and principal medical statistician with the Translational Epidemiology Unit, Nuffield Department of Population Health, Oxford University. Her research interest include at the intersection of machine learning and medical statistics, for both non-communicable and infectious diseases.
\end{IEEEbiography}

\begin{IEEEbiography}
[{\includegraphics[width=1in,height=1.25in,clip,keepaspectratio]{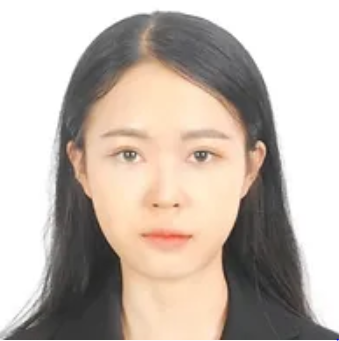}}]{Zijun Liu}
is a Ph.D. student in the Department of Biomedical Engineering, City University of Hong Kong, Hong Kong. 
\end{IEEEbiography}

\begin{IEEEbiography}
[{\includegraphics[width=1in,height=1.25in,clip,keepaspectratio]{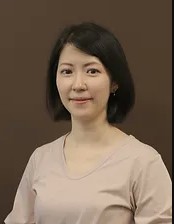}}]{Kannie. W.Y.  Chan} received her BSc and PhD degrees from The University of Hong Kong. She conducted post-doctoral research in biomaterials and medical imaging with a focus on MRI at Department of Radiology at Johns Hopkins University School of Medicine in 2010, and became the Assistant Professor in 2014. She joined The City University of Hong Kong in 2016. Her research focuses on the development of biomaterials and imaging techniques to facilitate the clinical translation of cell therapy and cancer therapy. This includes the use of an emerging MRI contrast mechanism for molecular imaging, which is known as chemical exchange saturation transfer (CEST). She published over 30 peer-reviewed articles, including the cover article in Nature Materials.
\end{IEEEbiography}

\begin{IEEEbiography}
[{\includegraphics[width=1in,height=1.25in,clip,keepaspectratio]{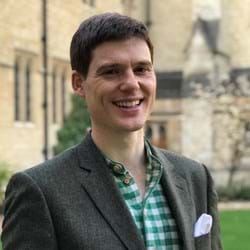}}]{David A. Clifton} is a Professor of Clinical Machine Learning and leads the Computational
Health Informatics (CHI) Lab in the Department
of Engineering Science at the University of Oxford.
\end{IEEEbiography}
\begin{IEEEbiography}[{\includegraphics[width=1in,height=1.25in,clip,keepaspectratio]{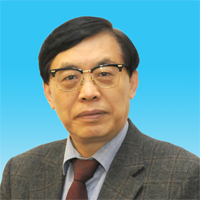}}]{Yuan-Ting Zhang}
is currently the Research Professor in the Department of Electronic Engineering at the Chinese University of Hong Kong, Visiting Professor at Oxford Suzhou Centre for Advanced Research (OSCAR), the MWLC-LRG member of Karolinska Institutet, Chief Scientist in unobtrusive blood pressure measuring technology at the Shenzhen Honor Mobile Terminal LtD, and the founding Chairman and first Director of Hong Kong Center for Cerebro-cardiovascular Health Engineering. He served as the Sensing System Architect in Health Technology and Sensing Hardware Divisions at Apple Inc., California, USA, the founding Director of the Key Lab for Health Informatics of Chinese Academy of Sciences (CAS), the founding Director of CAS-SIAT Institute of Biomedical and Health Engineering, Chair Professor at CityU, and Adjunct Chair Professor at Shandong University. Professor Zhang dedicated his service to the Chinese University of Hong Kong from 1994 to 2015, where he served as the first Head of the Division of Biomedical Engineering and led the efforts on establishing educational degree Programmes in Biomedical Engineering. Prof. Zhang has been the Editor-in-Chief for IEEE Reviews in Biomedical Engineering since 2016 and Chair of the IEEE 1708 Working Group. He was the Editor-in-Chief for IEEE T-ITB and the first Editor-in-Chief of IEEE J-BHI. He served as Vice Preside of IEEE EMBS. He was also the Chair of 2016-2018 IEEE Award Committee in Biomedical Engineering and a member of IEEE Medal Panel for Healthcare Technology Award. Prof. Zhang's research interests include unobtrusive sensing and wearable devices, and neural muscular modeling. He was selected on the lists of China’s Most Cited Researchers by Elsevier and the top 2\% researcher worldwide by Stanford University. He won a number of international awards including IEEE EMBS best journal paper awards, IEEE EMBS Outstanding Service Award, IEEE-SA 2014 Emerging Technology Award, and Earl Owen Lecture at SMIT-IBEC2018 in Korea. Prof. Zhang is elected as IAMBE Fellow, IEEE Fellow, AIMBE Fellow and AAIA Fellow for his contributions to the development of wearable and m-Health technologies.
\end{IEEEbiography}

\begin{IEEEbiography}
[{\includegraphics[width=1in,height=1.25in,clip,keepaspectratio]{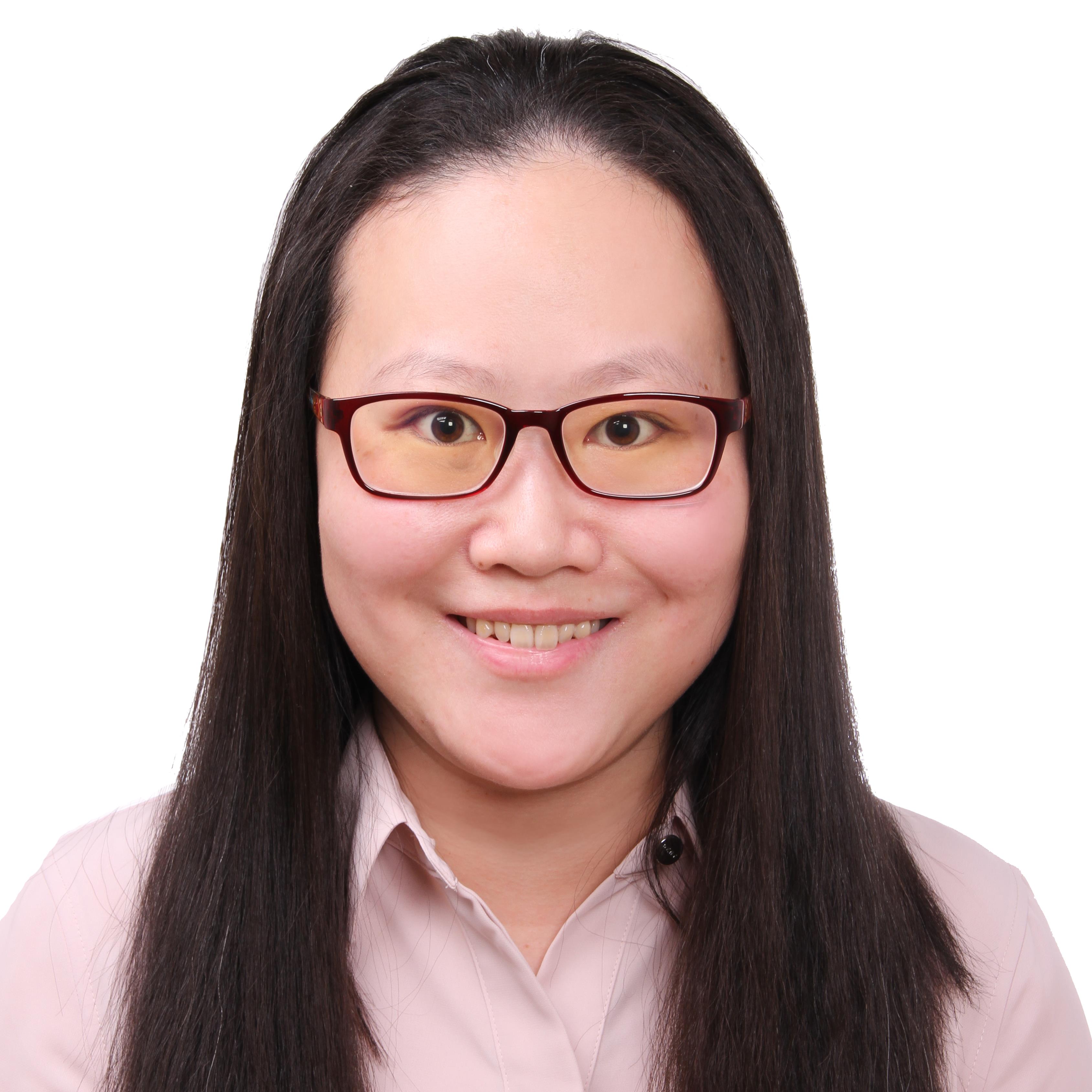}}]{Yining Dong}
received the B.Eng. degree in electronic engineering from Tsinghua University, Beijing, China, in 2011, and the Ph.D. degree in electrical engineering from University of Southern California, Los Angeles, CA, USA, in 2016, She is currently an Assistant Professor with the School of Data Science, City University of Hong Kong, Hong Kong. Her research interests include process data analytics, statistical machine learning, smart manufacturing, and new material design.
\end{IEEEbiography}



\end{document}